\documentclass[twoside]{article}

\usepackage[round]{natbib}     %
\usepackage[hidelinks]{hyperref}    %

\usepackage{graphicx} %
\usepackage{amsmath,amsfonts,bm,amssymb}

\usepackage{latexsym}
\usepackage{amssymb}
\usepackage{amsmath}
\usepackage{amsthm}
\usepackage{booktabs}
\usepackage{enumitem}
\usepackage{graphicx}
\usepackage{color}
\usepackage{graphicx}%
\usepackage{multirow}%
\usepackage{amsmath,amssymb,amsfonts}%
\usepackage{amsthm}%
\usepackage{mathrsfs}%
\usepackage[title]{appendix}%
\usepackage{xcolor}%
\usepackage{textcomp}%
\usepackage{manyfoot}%
\usepackage{booktabs}%
\usepackage{algorithm}%
\usepackage{algorithmicx}%
\usepackage{algpseudocode}%
\usepackage{listings}%
\usepackage{bm}%
\usepackage{todonotes}%
\usepackage[utf8]{inputenc}
\usepackage{mathtools}
\usepackage{yhmath}
\usepackage{multirow}

\usepackage{comment}

\newtheorem{theorem}{Theorem}[section]
\newtheorem{corollary}{Corollary}[theorem]
\newtheorem{definition}{Definition}[theorem]

\newtheorem*{problem}{Problem}

\newtheorem{example}[theorem]{Example}
\newtheorem{lemma}[theorem]{Lemma}

\newenvironment{proof*}[1][\proofname]{\proof[#1]}
{\endproof}

\newcommand{\bigO}{\mathcal{O}}

\newenvironment{continuedproof}[1]
  {\begin{proof}[Proof of #1 (continued)]}
  {\end{proof}}

\newenvironment{continuedproof*}[1]
  {%
   \begin{proof}[Proof of #1 (continued)]}
  {\end{proof}}

\usepackage{mdframed}

\newmdenv[
  skipabove=\topsep,
  skipbelow=\topsep,
  linewidth=0.8pt,
  linecolor=black,
  roundcorner=4pt,
  innertopmargin=0.8\baselineskip,
  innerbottommargin=0.8\baselineskip,
  innerleftmargin=10pt,
  innerrightmargin=10pt,
]{auxproofbox}

\DeclareMathOperator{\EX}{\mathbb{E}}%

\usepackage[accepted]{aistats2026}

\begin{document}

\twocolumn[

\aistatstitle{Information Hidden in Gradients %
of  %
Regression with Target Noise}

\aistatsauthor{ Arash Jamshidi \And Katsiaryna Haitsiukevich \And  Kai Puolamäki}

\aistatsaddress{ University of Helsinki \And  University of Helsinki \And University of Helsinki} ]

\begin{abstract}
Second-order information---such as curvature or data covariance---is critical for 
optimisation, diagnostics, and robustness. However, in many modern settings, only the gradients are observable. We show that the gradients alone can reveal the Hessian, equalling the data 
covariance $\Sigma$ for the linear regression. Our key insight is a simple variance calibration: injecting Gaussian noise so that the total target noise variance equals the batch size ensures that the empirical gradient covariance closely approximates the Hessian, even when evaluated far from the optimum. We provide non-asymptotic operator-norm guarantees under sub-Gaussian inputs. We also show that without such calibration, recovery can fail by an $\Omega(1)$ factor. 
The proposed method is practical (a ``set target-noise variance to $n$’’ rule) and 
robust (variance $\mathcal{O}(n)$ suffices to recover $\Sigma$ up to scale). Applications 
include preconditioning for faster optimisation, adversarial risk estimation, and
gradient-only training, for example, in distributed systems. We support our theoretical results with experiments on synthetic and real 
data.
\end{abstract}

\section{Introduction}

Gradients are fundamental in modern machine learning. They drive optimisation algorithms such as stochastic gradient descent (SGD), yet their usefulness extends far beyond optimisation. Gradients also serve as statistical objects that reveal structural information about the loss and the underlying data distribution. For instance, Hessian–vector products can be approximated using gradients at nearby points (\cite{pearlmutter1994fast}), providing a practical way to estimate curvature, and gradient-based stopping rules have been proposed to improve generalisation (\cite{jamshidi2025,mahsereci2017early,forouzesh2021disparity}). These examples illustrate that gradients encode rich information that can be extracted in addition to their role in optimisation.  

In this work, we investigate whether gradients can reveal the Hessian of the loss function.  We consider the setting where the learner does not observe raw data but only average gradients over batches of data. This situation naturally arises in distributed and federated learning, where sharing data is infeasible but gradients are communicated. Another use case is the efficient implementation of optimisation methods in machine learning, where the optimisation algorithm may have easy access to only gradient information. 
Formally, our focus is on linear regression,  
\begin{align}
y = x^\top w_0 + \varepsilon, \quad x = \varphi(z), \quad (x,y)\in\mathbb{R}^d \times \mathbb{R},
\end{align}
such that $z$ is the data point and $\varphi$ is a possible non-linear transformation of $z$. We assume covariates $x \sim \mathcal{D}_x$ satisfying $\mathbb{E}[x]=0$ and $\mathbb{E}[xx^\top]=\Sigma \preceq I$, and target noise $\varepsilon \sim \mathcal{N}(0,\widetilde{\sigma}^2)$. We assume access only to aggregate batch gradients of the squared loss, for any batch $B=\{(x_i,y_i)\}_{i=1}^n$.  
We observe $k$ such independent batches. Our goal is to recover the Hessian $\Sigma$ using only these gradients. Later, in the experiments, we show that the results hold for non-linear models.

\begin{problem}[Estimating $\Sigma$ from gradients]\label{main:problem}
Given $\epsilon>0$, construct an estimator $\widehat{\Sigma}$ from batch gradients such that, with high probability,
\begin{align}
\|\widehat{\Sigma}-\Sigma\|_{\mathrm{op}} \leq \epsilon,
\end{align}
where $\|\cdot\|_{\mathrm{op}}$ denotes the operator norm.
\end{problem}
This task is challenging because gradients are entangled with both the current 
parameter $w$ and the target noise $\varepsilon$. To address 
this, we highlight two key insights:  
\paragraph{Insight 1: Bias--Variance Calibration.}  
For sufficiently large batch size $n$, if we add artificial noise 
$\mathcal{N}(0,O(n))$ to the targets $y$, the gradient covariance 
reveals the structure of $\Sigma$:  
\begin{align}
    \mathrm{Cov}(\nabla L^{(B)}(w)) = \mathcal{O}(1)\,\Sigma + \Delta(w),
\end{align}
such that the nuisance term $\|\Delta(w)\|_\mathrm{op} \leq \epsilon$ and with the leading term proportional to $\Sigma$.\footnote{We use the usual definition of covariance of a random vector Y, $\mathrm{Cov}(Y) = \mathbb{E}[(Y - \mathbb{E}(Y))(Y - \mathbb{E}(Y))^\top]$.}

\paragraph{Insight 2: Noise does not corrupt mean gradients.}  
Injecting $\mathcal{N}(0,O(n))$ noise into the targets does not significantly alter the mean gradient across batches. Specifically, by defining $\nabla L^*(w) := \Sigma(w-w_0)$ as the true (population-level) gradient, for $k = \Omega(d/\epsilon^2)$, the noisy and true gradients satisfy  
\begin{align}
    \left\|\frac{1}{k}\sum\nolimits_{j=1}^k \nabla L^{(j)}(w) 
    - \nabla L^*(w)\right\|_2 \leq \epsilon
\end{align}
with high probability. 

\begin{figure}[t]
    \centering
    \includegraphics[width=\linewidth]{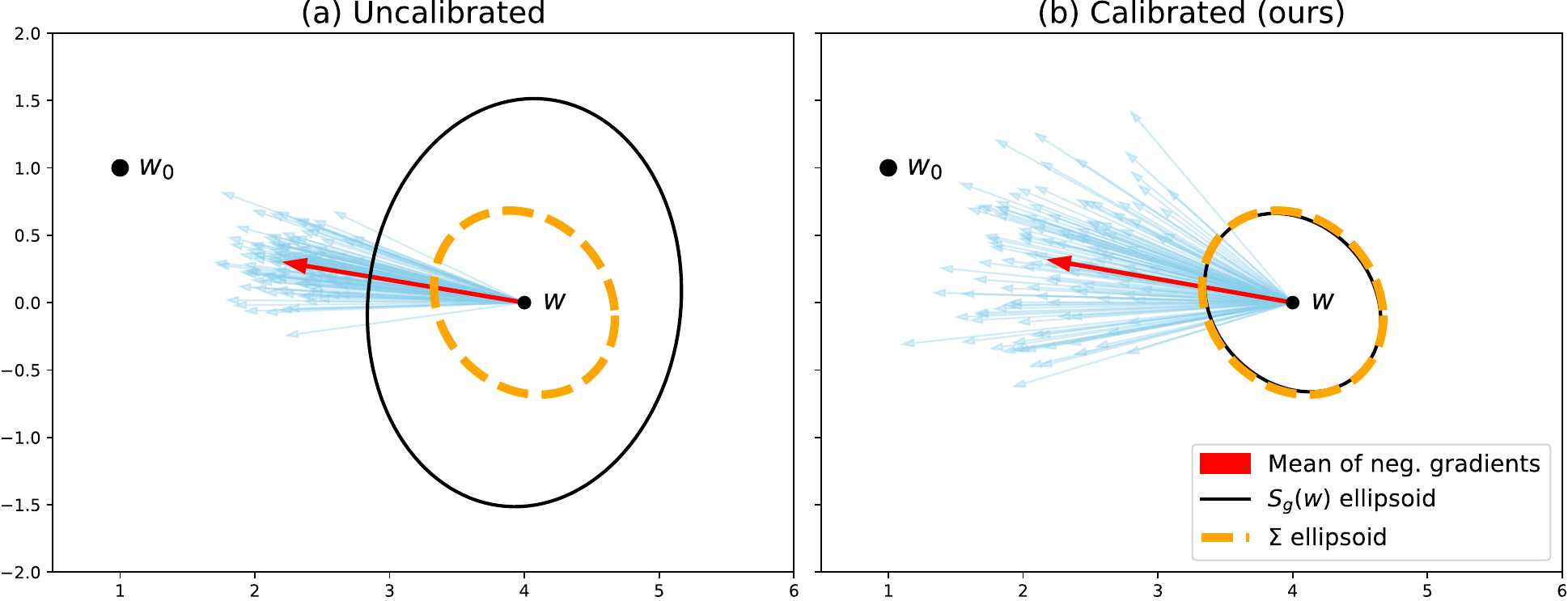}
\caption{%
\textbf{Gradient covariance calibrated with target noise matches the Hessian (Insight 1) and preserves the mean gradient (Insight 2).}
Blue thin lines: batch gradients; red thick line: their average; black solid ellipse: empirical covariance $S_g(w)$; 
orange dashed ellipse: data covariance $\Sigma$. \textit{Left}: without calibration, $S_g(w)$ 
deviates from $\Sigma$. \textit{Right}: after adding noise to the targets, the gradients 
spread out and $S_g(w)$ aligns closely with $\Sigma$, while the mean gradient 
remains accurate.}
    \label{fig:intro}
\end{figure}

Thus, noise calibration \emph{preserves the accuracy of the gradient signal} 
while \emph{enabling accurate recovery of $\Sigma$}. These two insights are summarised in Fig.~\ref{fig:intro}. 

\paragraph{Algorithmic Idea.}  
Guided by these insights, we inject Gaussian noise of variance $O(n)$ into the 
targets $y$ and then use the empirical gradient covariance of the batches $S_g(w)$ as an 
estimator of $\Sigma$.

\paragraph{Our Contributions.}  
This work makes four main contributions:  
\textbf{(1)} We formalize the problem of estimating the Hessian $\Sigma$ from 
gradient information alone. The gradient covariance encodes $\Sigma$ but is 
generally biased. We provide a non-asymptotic analysis showing that, by injecting Gaussian noise into the targets at variance $O(n)$ and using the empirical gradient 
covariance $S_g(w)$, one can consistently recover $\Sigma$ for any sub-Gaussian distribution $\mathcal{D}_x$ under mild assumptions, with high probability.
  \textbf{(2)} We show that \emph{noise injection is not only sufficient but also necessary}. 
    Without calibration, the operator-norm distance between gradient covariances and 
    $\Sigma$ can remain $\Omega(1)$, highlighting that adding 
    Gaussian noise at variance $O(n)$ with a sufficient batch size is essential for recovery.  
\textbf{(3)} We demonstrate practical implications in gradient-only settings (e.g., efficient implementation of optimisation algorithms with access only to the gradient information, occurring, for example, in distributed systems): 
(i) estimating adversarial risk when only gradient access is available, and
    (ii) using calibrated gradient covariance for preconditioning, thereby accelerating 
    convergence.
\textbf{(4)} We show the effectiveness of our method in estimating the Hessian, in both linear and non-linear (ReLU neural networks) models, using real-world and synthetic data. Our code is available at \url{https://github.com/edahelsinki/noisy_targets}.

\section{Related Work}
The addition of purposeful noise during model training can be considered from several angles.

\textbf{Noise injection as a regulariser}
Noise injection is a common regularisation method, introduced via sampling, such as dropout \citep{srivastava2014dropout} or SGD \citep{robbins1951stochastic,bottou1991stochastic}, or via additive perturbations \citep{dhifallah2021inherent}. It can be applied to model inputs \citep{dhifallah2021inherent,cohen2019certified}, layer inputs \citep{orvieto2023explicit}, weights \citep{camuto2020explicit,wu2020adversarial}, or gradients \citep{zhu2018anisotropic,wen2020empirical,wu2020onthenoisy}, with gradient noise particularly effective for generalisation and large-batch training \citep{wu2020onthenoisy,zhu2018anisotropic,wen2020empirical}. Our approach---adding noise to targets---is closest to gradient perturbation, but introduces noise to the gradients implicitly and recovers the Hessian, enabling faster convergence when used as a preconditioner.

\textbf{Hessian estimate by gradients using Fisher Information}
Second-order optimisation methods often use the Fisher Information matrix as a surrogate for the Hessian \citep{martens2020new,sen2024sofim,jhunjhunwala2024fedfisher}. Since it is difficult to compute directly, it is typically approximated by the empirical Fisher, i.e., the unnormalised uncentered gradient covariance \citep{rame2022fishr}. However, this approximation may not be reliable since its underlying assumptions rarely hold in practice \citep{kunstner2019limitations}. Additionally, the empirical Fisher information matrix uses point-wise gradients calculated from the model outputs, which can be computationally expensive to obtain, especially if the size of the output is large. In contrast, our method uses the batch gradient of the loss, which is much more computationally efficient. We further demonstrate that the equality between the noise-free gradient covariance and the Hessian does not hold, even in linear regression.

\textbf{Quasi-quadratic optimisation}
This class of optimisation methods estimates the second-order information from the gradients \citep{martens2010deep,byrd2016stochastic,goldfarb2020practical,berahas2022quasi}.
The main difference between our method and quasi-quadratic approaches is that quasi-quadratic methods focus on optimisation problems and building an efficient optimiser, while we use the statistical properties of the problem to extract information about the Hessian in general, with several possible applications, optimisation being one of them. Additionally, with our method, the weight updates can be performed with the mean of the noisy gradients to avoid calculating the forward and backward paths twice and more importantly, to reduce the information that is shared between nodes in case of distributed training.

\textbf{Adversarial attacks and defences}
Adversarial attacks often involve adding carefully crafted noise to inputs to alter model predictions \citep{szegedy2013intriguing,kurakin2018adversarial,xu2023best,kong2025adversarial}. In contrast, our noise injection is not learned and does not target prediction changes; instead, we preselect a suitable Gaussian noise level for targets to estimate the Hessian structure.
On the other hand, federated learning trains models collaboratively by sharing parameter updates or gradients while keeping data local. Prior work shows that gradients can leak information about the data distribution, such as the target distribution \citep{wainakh2021user,kariyappa2023exploit}. A common defence is adding noise to gradients \citep{li2024staged,wan2023enhancing}. We show that carefully designed noise can also reveal the covariance structure of the inputs.

\section{Warm-up: Equal Hessian Losses}
Before analysing the link between gradient covariance and the Hessian 
(or equivalently, the data covariance) in linear regression, we consider 
a simplified case. Let $L^{(i)}(w)$, $\nabla L^{(i)}(w)$, and 
$\nabla^2 L^{(i)}(w)$ denote the loss, gradient, and Hessian for batch $i$. 
Assume each $L^{(i)}$ is a quadratic function with shared Hessian 
$\nabla^2 L^{(i)} = \Sigma$ for all $i \in [k]=\{1,\ldots,k\}$. 
In this setting, we have:
\begin{example}\label{thm:equalhessian}
Let each batch $i \in [k]$ be associated with a general quadratic loss function of the form
\begin{align}
L^{(i)}(w) = a_i + b_i^{\top} w + w^{\top} \Sigma w/2,
\end{align}
where $a_i \in \mathbb{R}$, $b_i \in \mathbb{R}^d$, and $(a_i, b_i) \overset{\text{i.i.d.}}{\sim} \mathcal{D}$ for some unknown distribution $\mathcal{D}$. Suppose $\Sigma \in \mathbb{R}^{d \times d}$ is a fixed, symmetric, positive definite matrix. Define
\begin{align}
w_*^{(i)} := \arg\min_{w \in \mathbb{R}^d} L^{(i)}(w), \quad 
\widehat{w} :=  \sum\nolimits_{i=1}^k w_*^{(i)}/k.
\end{align}
Then, for $w \in \mathbb{R}^d$, the gradient covariance is given by
\begin{align}
S_g(w) = \Sigma \widehat{S}_{w_*} \Sigma,
\end{align}
where $\widehat{S}_{w_*} := \frac{1}{k} \sum\nolimits_{i=1}^k (w_*^{(i)} - \widehat{w})(w_*^{(i)} - \widehat{w})^{\top}$ is the empirical covariance of the optimal points $\{w_*^{(i)}\}_{i=1}^k$.
\end{example}

While $\Sigma$ may differ between batches in practice (e.g., due to finite data effects), the central intuition in this chapter is that by controlling the distribution of $\widehat{S}_{w_*}$, 
we can alter the distribution of $S_g(w)$. In the case of linear regression, one way to achieve this is by manipulating the targets, for example, by adding noise.

\section{Gradient Covariance Recovers $\Sigma$ with Noise Injection}\label{sec:non-asymp}
In this section, we show that the gradient covariance $S_g(w)$ approximates the Hessian 
$\Sigma$ in linear regression when the target noise is $\mathcal{N}(0,n)$, with $n$ the 
batch size, under both large- and small-batch regimes. Note that if the original targets already have target noise with known variance $\widetilde{\sigma}^2 < n$, 
it suffices to inject additional noise $\mathcal{N}(0,\,n-\widetilde{\sigma}^2)$ so that the total noise variance is $n$. 
We also discuss this case, as well as when $\widetilde{\sigma}^2$ is unknown, later, 
in Section~\ref{apps:federated}.

In both regimes, we assume 
$n = \Omega(\|w-w_0\|_2^2)$ and $k = \Omega(d/\epsilon^2)$, which ensures the batch 
gradients are sub-exponential $SE_d(\mathcal{O}(1),\mathcal{O}(1))$ (see Appendix~\ref{theory:subexp}). Consequently, the average of $k$ batches concentrates around the population gradient:
\begin{align}
    \Bigg\|\frac{1}{k}\sum\nolimits_{j=1}^k \nabla L^{(j)}(w) - 
    \Sigma(w-w_0) \Bigg\|_2 
    \lesssim \epsilon, \quad \epsilon < 1,
\end{align}
by standard concentration bounds  \citep{wainwright2019high}. 
Thus, while recovering $\Sigma$, we also preserve accurate gradient estimation required 
for optimisation.
\subsection{Large Batch Size}\label{sec:largebatch}
We prove our main result (gradient covariance estimates Hessian with target noise $\sigma^2=n$) when the batch size is large.
\begin{theorem} \label{thm:main}
    Let \( j \in [k] \) index a batch of size \( n \), and suppose the inputs \( x_i^{(j)} \overset{\text{i.i.d.}}{\sim} \mathcal{D}_x\) such that $\mathcal{D}_x \in SG_d(1)$ (sub-Gaussian) such that $\mathbb{E}[x] = 0$ and $\mathbb{E}[x x^\top] = \Sigma \preceq I$, the noise terms \( \varepsilon_i^{(j)} \overset{\text{i.i.d.}}{\sim} \mathcal{N}(0, \sigma^2 = n) \), and the targets are given by \( y_i^{(j)} = (x_i^{(j)})^\top w_0 + \varepsilon_i^{(j)} \), for $i \in [n]$. Define the empirical least-squares loss over batch \( j \) as
\begin{align}\label{eq:l2loss}
L^{(j)}(w) = \frac{1}{2n} \|X^{(j)}w - y^{(j)}\|_2^2,
\end{align}
and let \( \nabla L^{(j)}(w) \) denote the gradient of the loss with respect to \( w \). Then, by defining the average gradient
\begin{align}
    \nabla L := \frac{1}{k} \sum\nolimits_{\ell=1}^k \nabla L^{(\ell)}(w),
\end{align}
the empirical batch gradient covariance is given by
\begin{align}
    S_g(w) = \frac{1}{k} \sum_{j=1}^k \left( \nabla L^{(j)}(w) - \nabla L \right) \left( \nabla L^{(j)}(w) - \nabla L \right)^\top. \nonumber
\end{align}
Then, by defining $c := \lambda_{min}(\Sigma)$ as a constant, there exists dimension independent absolute constants $C_1, C_2, C_3$, and $C_4$ such that if batch size $n$ and number of batches $k$ satisfy the following:
\begin{align}
    &n \geq C_2 d^2(d+\log(k/\delta))/
    \epsilon^2\\
    &d \geq C_4[\log(2k) + \log(1/\delta)] \\
    &k \geq C_1(d + \log(1/\delta))/\epsilon^2,
\end{align}
then, with probability at least $1 - 4\delta$, we have 
\begin{align}
    \|S_g(w) - \Sigma\|_\mathrm{op} \leq \epsilon
\end{align}
for all $\{w :||w - w_0|| \leq C_3\sqrt{d}\}$.
\end{theorem}

The proof builds on Example~\ref{thm:equalhessian}, using the fact that for large 
batch size the Hessians of different batches are, with high probability, close in 
operator norm. The argument then follows from standard concentration and covering techniques.

We note two points about Theorem~\ref{thm:main} and its proof:

(i) Although Theorem~\ref{thm:main} is stated assuming equal batch size $n$, it is only for notational simplicity. The proof extends to unequal sizes 
$n_1,\dots,n_k$, provided
\begin{align}
\min(n_1, n_2, \dots, n_k) \;\;\geq\;\; C_2 d^2(d+\log(k/\delta))/
    \epsilon^2.
\end{align}
(ii) A novel aspect of our analysis is that for sufficiently large batch size $n$, we guarantee
\begin{align}
\|S_g(w) - \Sigma\|_{\mathrm{op}} \lesssim \epsilon
\end{align}
uniformly for all $w$ near $w_0$, using only $\Theta(d/\epsilon^2)$ batches---the sharp complexity already needed for concentration at a single point. Typically, uniform guarantees 
require far more batches; our improvement comes from exploiting correlations in large batches 
within a refined covering argument, avoiding extra logarithmic or polynomial factors in $k$.

\subsection{Small Batch Size}
Section~\ref{sec:largebatch} showed recovery of $\Sigma$ from $S_g(w)$ with noisy 
targets when batches are large. We now argue it also holds for smaller batches, 
for any $w$, provided $k = \widetilde{\Omega}(d/\epsilon^2)$.\footnote{See Appendix \ref{app:notation} for the definition of $\widetilde{\Omega}$ and other notations.} 
We begin with the following lemma:
\begin{lemma}\label{lemma:smallbatch}
    Let \( j \in [k] \) index a batch of size \( n \), and suppose the inputs \( x_i^{(j)} \overset{\text{i.i.d.}}{\sim} \mathcal{D}_x \), such that $\mathcal{D}_x \in SG(1)$ with $\mathbb{E}[x] = 0$, $\Sigma= \mathbb{E}[xx^\top] \preceq I$, the noise terms \( \varepsilon_i^{(j)} \overset{\text{i.i.d.}}{\sim} \mathcal{N}(0, \sigma^2) \), and the targets are given by \( y_i^{(j)} = (x_i^{(j)})^\top w_0 + \varepsilon_i^{(j)} \), for $i \in [n]$. Define the empirical least-squares loss over batch \( j \) as $L^{(j)}(w)$ (as in Equation~\eqref{eq:l2loss})
and let \( \nabla L^{(j)}(w) \) denote the gradient of the loss with respect to \( w \). Then we have:
\begin{align}
\mathrm{Cov}(\nabla L^{(j)}(w)) &= \frac{\sigma^2}{n} \Sigma + \frac{\mathrm{noise}}{n},\\
\text{s.t. }    \|\mathrm{noise}\|_\mathrm{op} &\leq C \|w - w_0\|_2^2,
\end{align}
for some absolute constant $C$.
\end{lemma}
From Lemma~\ref{lemma:smallbatch}, for any fixed $w$, if $\sigma^2 = n$ and the 
batch size satisfies $n = \Omega(\|w-w_0\|_2^2/\epsilon)$, then
\begin{align}
    \|\mathrm{Cov}(\nabla L^{(j)}(w)) - \Sigma\|_\mathrm{op} \leq \epsilon,
\end{align}
as the operator norm of the nuisance term scales linearly with batch size $n$, and we proved $\|\mathrm{noise}\|_\mathrm{op} \leq C \|w - w_0\|_2^2$. So, the intuition behind our method is that by increasing the batch size and $\sigma^2$ at the same time, we preserve $\Sigma$ while forcing the nuisance terms to become smaller. Moreover, because batch gradients are i.i.d. 
sub-exponential random vectors, $SE_d(\mathcal{O}(1),\mathcal{O}(1))$, using any covariance estimator $\widehat{S}_g(w)$ suitable for sub-exponential 
vectors (e.g., with truncation \citep{ke2019user}) and $k = \widetilde{\Omega}(d/\epsilon^2)$ 
batches, we obtain
\begin{align}
    \|\widehat{S}_g(w) - \Sigma\|_\mathrm{op} \lesssim \epsilon
\end{align}
with high probability. We do not elaborate further, as this is a direct application of the standard 
covariance estimation for sub-exponential vectors. Note that by using only these properties, a uniform convergence similar to Section~\ref{sec:largebatch} requires more batches, by the union bound.

\section{Recovery of $\Sigma$ Fails without Calibration}\label{sec:expected}
While the analysis in Section~\ref{sec:non-asymp} demonstrates that when the target noise variance matches the batch size (i.e., \( \sigma^2 = n \)), the empirical gradient covariance \( S_g(w) \) closely approximates the population covariance \( \Sigma \), it remains unclear from the non-asymptotic perspective whether adding artificial noise to the target variable \( y \) is actually necessary.

In this section, we answer this question affirmatively by analysing the exact (population-level) covariance of the gradients. We show that if the data points are i.i.d.\ draws from a multivariate Gaussian distribution, \( x_i^{(j)} \overset{\text{i.i.d.}}{\sim} \mathcal{N}(0, \Sigma) \), for all \( i \in [n] \), \( j \in [k] \), with \( \Sigma \succ 0 \), then the choice \( \sigma^2 = \mathcal{O}(n) \) is necessary to recover \( \Sigma \) from \( S_g(w) \).

\begin{lemma}\label{thm:expected}
Under the setting of Lemma~\ref{lemma:smallbatch}, if we have  \( x_i^{(j)} \overset{\text{i.i.d.}}{\sim} \mathcal{N}(0, \Sigma) \), then we obtain:
\begin{align}
\mathrm{Cov}(\nabla L^{(j)}(w)) &= (\sigma^2/n)\, \Sigma 
\nonumber\\
&+   \Sigma(w - w_0)(w - w_0)^\top \Sigma /n
\nonumber\\
&+ ((w - w_0)^\top \Sigma (w - w_0))  \Sigma/n .
\end{align}
\end{lemma}
Hence, by Lemma~\ref{thm:expected}, we show that recovery of $\Sigma$ fails without noise injection:
\begin{corollary}\label{cor:necessary}
    Under the setting of Lemma~\ref{thm:expected}, for any constants noise variance $\sigma^2$ and batch size $n$, there exists $w \in \mathbb{R}^d$ such that for any $j \in [k]$ we have 
    \begin{align}
            \|\mathrm{Cov}(\nabla L^{(j)}(w)) - \Sigma\|_{\mathrm{op}}/ \|\Sigma\|_\mathrm{op} \geq 0.99.
    \end{align}
\end{corollary}
To further elaborate on the importance of Lemma~\ref{thm:expected}, we show that the worst-case behaviour could not simply be derived using the general result of Lemma~\ref{lemma:smallbatch}:
\begin{lemma}\label{lem:ctrexample} In dimension $d = 1$, there exists a (non-Gaussian) distribution $\mathcal{D}_x$ with $\mathbb{E}[x] = 0$ and $\mathbb{E}[x^2] = \Sigma = 1$, $w_0 \in \mathbb{R}$, such that for any $w \in \mathbb{R}$, batch size $n$, we have:
\begin{align}
    \mathrm{Cov}(\nabla L^{(j)}(w)) =(\sigma^2/n)\, \Sigma.
\end{align}
Hence, as long as $\sigma^2 = n$, we have $\mathrm{Cov}(\nabla L^{(j)}(w))=\Sigma$, for all $w \in \mathbb{R}^d$.
\end{lemma}
\section{Gradient-Only Applications}\label{apps:federated}
In many modern machine learning scenarios, a training algorithm receives only gradients while raw data remains hidden. Such situations also arise in federated or distributed learning 
(\cite{mcmahan2017communication, kairouz2021advances}), as well as in privacy-sensitive 
applications and bandwidth-limited systems. In such settings, the statistical properties of the gradients become the key source of information.  

Our analysis enables estimation of the population covariance $\Sigma$ from gradients 
alone, making it broadly applicable in gradient-only regimes. We highlight two 
applications: (i) designing preconditioners for faster convergence, 
since $\Sigma$ coincides with the Hessian of the loss; and (ii) estimating adversarial 
risk from gradients, enabling robustness assessment without data access. 
Algorithm~\ref{alg:central} summarises these ideas.

\begin{algorithm}[t]
\caption{Optimisation with Noisy Gradients\label{alg:central}. At each step $t$, $\|w_t-w_0\|^2_{\Sigma}$ can be estimated by $\nabla L(w_t)^\top (S_g(w_t))^{-1} \nabla L(w_t)$ (Lemma \ref{lemma:advrisk}) and $E^{\|\cdot\|}(w_t, w_0, \beta)$ by Lemma \ref{lemma:advriskref}.}
\begin{algorithmic}[1]
\Require Starting parameter $w_1 \in \mathbb{R}^d$, $\#$ epochs $T$,  $\#$ batches $k$, batch size $n$, inherent target noise estimate $\widetilde{\sigma^2}$, target values $y^{(j)} \in \mathbb{R}^{n}$ and gradient function $\nabla L^{(j)}(w,y)\in\mathbb{R}^d$ for every batch $j \in [k]$, optimisation update $\mathrm{OPTIMISER} : (w_t, g_t) \mapsto w_{t+1}$ 
where $w_t \in \mathbb{R}^d$ is the current parameter 
and $g_t \in \mathbb{R}^d$ is the preconditioned gradient.

\State $y^{(j)} \leftarrow y^{(j)} + \mathcal{N}\left(\overline 0, (n - \widetilde{\sigma^2})I_{n \times n}\right)$ for all $j\in[k]$ %
\For{$t = 1$ to $T$} \Comment{Number of epochs}
\State $\mathbf{G}_t = \{\nabla L^{(j)}(w,y^{(j)})\}_{j\in[k]}$ \Comment{Noisy gradients }
\State $\nabla L(w_t) \leftarrow \mathrm{Mean}(\mathbf{G}_t)$ \Comment{Gradient mean}
\State $S_g(w_t) \leftarrow \mathrm{Cov}(\mathbf{G}_t)$ 
\Comment{Gradient covariance}
\State %
\hyperref[thm:faster-converge]{
    $w_{t+1} \gets \mathrm{OPTIMISER}\left(w_t, (S_g(w_t))^{-1}\nabla L(w_t)\right)$}\Comment{See Sect. \ref{sec:inversion} for efficient matrix inversion}
\EndFor
\end{algorithmic}
\end{algorithm}

We assume targets follow $ y = x^\top w_0 + \varepsilon $ such that 
$\varepsilon \sim \mathcal{N}(0,\widetilde{\sigma}^2)$ and 
$\widetilde{\sigma}^2 < n$. To align gradient covariance $S_g(w)$ with $\Sigma$, 
we add independent noise $\mathcal{N}(0,\,n-\widetilde{\sigma}^2)$, so the total variance is equal to $n$.  In practice, $\widetilde{\sigma}^2$ may be unknown. If $\widetilde{\sigma}^2 < n$, 
adding noise $\mathcal{N}(0,n)$ yields
\begin{align}
S_g(w) \approx c\Sigma, \quad c \in [1,2],
\end{align}
recovering $\Sigma$ up to scale (by Lemma~\ref{lemma:smallbatch}), sufficing 
for most applications, where only the relative geometry of $\Sigma$ matters. We also assumed $\mathbb{E}[x] = 0$ and no bias term. In practice, this can be corrected by centring $x$ and $y$ in each batch using their empirical means.
\subsection{Preconditioning and Faster Convergence}\label{thm:faster-converge}
In this section, we show that we can accelerate the convergence rate of gradient descent for the problem of linear regression by estimating the Hessian $\Sigma$ using $S_g(w)$. The standard convergence result of gradient descent with constant step size is that to get error $\eta$ we need $\approx \Omega(\kappa \log{1/\eta})$ steps (\cite{nesterov2013introductory}), where $\kappa = \lambda_{\mathrm{max}}/\lambda_{\mathrm{min}}$, is the condition number.
The above convergence rate is particularly slow when the condition number $\kappa$ is large, but can be fixed using a good preconditioner, e.g., $(S_g(w))^{-1}$, such that $\|(S_g(w))^{-1} - \Sigma^{-1}\|_\mathrm{op} \leq \epsilon$.
\begin{theorem}\label{theorem:precond}
It is possible to significantly accelerate the convergence rate of gradient descent in linear regression, using noisy gradient covariance $S_g(w)$ if for each $w$ such that $\|w - w_*\|_2 < \|w_1 - w_*\|_2$, we have  $\|(S_g(w))^{-1} - \Sigma^{-1}\|_\mathrm{op} \leq \epsilon$, where $w_1$ is the starting parameter. To achieve an error of $\eta$, one can reduce the number of required steps from
\begin{align}
    \Omega(\kappa \log{(1/\eta)}) \quad \text{to} \quad \Omega\left(\frac{\log (1/\eta)}{\log (1/\epsilon)}\right)
\end{align}
where $\kappa = \lambda_{\mathrm{max}}/\lambda_{\mathrm{min}}$ is the condition number of Hessian (data covariance) $\Sigma \preceq I$. This demonstrates a method to achieve a condition-number-independent convergence rate.
\end{theorem}
\subsection{Estimating Adversarial Risk}
In this section, we show that the test-time adversarial risk in linear regression 
can be estimated from gradients alone, provided sufficient noise is added to the targets. This result also offers a principled way to perform 
early stopping using only gradient information \citep{scetbon2023robust, javanmard2022precise, xing2021adversarially}.

\begin{lemma}[Informal]\label{corr:adv} It is possible to estimate the adversarial risk $E^{\|\cdot\|}(w, w_0, \beta)$ using only noisy gradient information, for every $w$ in a neighborhood of $w_0$.
\end{lemma}
To formalise this, we begin with the following definition, adapted from \cite{scetbon2023robust, madry2017towards, xing2021adversarially}.
\begin{definition}[Adversarial Risk]\label{def:advrisk}
Let \( w \in \mathbb{R}^d \) and \( \beta \geq 0 \). The adversarial risk of \( w \) at radius \( \beta \) is defined as
\begin{align}
E^{\|\cdot\|}(w, w_0, \beta) := \mathbb{E}_x \left[ \sup_{\|\delta\|_2 \leq \beta} \left( (x + \delta)^\top w - x^\top w_0 \right)^2 \right], \nonumber
\end{align}
where \( x \sim \mathcal{N}(0, \Sigma) \) is a test-time input sampled from the data distribution. When $\beta = 0$, we call $E^{\|\cdot\|}(w, w_0, 0)$ the Ordinary Risk.
\end{definition}
The intuition behind Definition~\ref{def:advrisk} is that each training example 
$x_i$ comes from $\mathcal{N}(0,\Sigma)$, and a test input $x$ may be perturbed 
by a worst-case noise vector $\delta$ with $\|\delta\|\leq \beta$. This models the standard adversarial setting, where inputs are corrupted within a norm ball to maximise error. The resulting risk measure can also serve as a conservative criterion for early stopping.

In linear regression with Gaussian data, the adversarial risk admits a closed-form expression:
\begin{lemma}[\citep{scetbon2023robust} Closed-Form Expression for Adversarial Risk]\label{lemma:advriskref} 
Let \( c_0 := \sqrt{2/\pi} \). Then, for any \( w \in \mathbb{R}^d \) and \( \beta \geq 0 \), the adversarial risk is given by
\begin{align}
E^{\|\cdot\|}(w, w_0, \beta) 
&= \|w - w_0\|_{\Sigma}^2 
\nonumber\\
&+ \beta^2 \|w\|_2^2 + 2 c_0 \beta \|w - w_0\|_{\Sigma} \|w\|_2. \nonumber
\end{align}
\end{lemma}
At first glance, Lemma~\ref{lemma:advriskref} seems to require knowing both $w_0$ 
and $\Sigma$. We show that this dependence can be avoided: adversarial risk can be 
estimated using only $w$ and noisy gradients.
\begin{lemma}[Quadratic Form Approximation of $\|w - w_0\|^2_\Sigma$ ]\label{lemma:advrisk} 
Let $E^{\|\cdot\|}(w, w_0, 0) = \|w - w_0\|^2_{\Sigma} = \bigO(1)$. Suppose we are given  gradient and gradient covariance $\nabla L(w) , S_g(w),$ such that
\begin{align}
\left\| \nabla L(w) - \Sigma(w-w_0) \right\|_2 &\leq \mathcal{O}(\epsilon), \\
\| \left(S_g(w)\right)^{-1} - \Sigma^{-1} \|_{\mathrm{op}} &\leq \mathcal{O}(\epsilon)
\end{align}
for some $\epsilon < 1$. Then, for some quadratic function based only on gradients, we can provide an estimate of \( \| w-w_0 \|^2_\Sigma \) with additive error \( \mathcal{O}(\epsilon) \); that is,
\begin{align}
\left| \nabla L(w)^\top 
\left( S_g(w) \right)^{-1} 
\nabla L(w)
- \|w - w_0\|^2_\Sigma \right| \leq \mathcal{O}(\epsilon). \nonumber
\end{align}
\end{lemma}
With an approximation of $\|w-w_0\|_\Sigma^2$, it is now easy to approximate $E^{\|\cdot\|}(w, w_0, \beta)$ (using Lemma~\ref{lemma:advriskref}), and hence establishing Lemma~\ref{corr:adv}. Accurate estimation of $E^{\|\cdot\|}(w, w_0, \beta)$ enables principled early stopping 
of optimisation algorithms using only gradient information. %

\section{Experiments}
\label{sec:experiments}
   
This section presents evaluations of the proposed method, along with its sensitivity analysis. We verify our theoretical results first on generated data from a Gaussian distribution and then extend the analysis to four real-world datasets. We test (a) linear regression models for the estimates of the data covariance (which is equal to the Hessian for linear models) and (b) multi-layer perceptrons (MLP) with ReLU activations for Hessian approximation. Both models are evaluated in the presence of added target noise and without it. We demonstrate the application of the Hessian estimate for adversarial risk calculation and as a preconditioner. 

In this section, the quality of the Hessian approximations is evaluated by a relative operator norm $r$ of the difference between $S_g(w)$ and $\Sigma$. The norm $r$ is calculated as follows:
\begin{equation}
    r = ||S_g(w) - \Sigma||_\mathrm{op}/||\Sigma||_\mathrm{op}
\label{eq:norm_r}
\end{equation}

\subsection{Method analysis with generated data}
\label{sec:exp_gen_data}

\begin{figure}[!t]
    \centering
    \includegraphics[width=\linewidth]{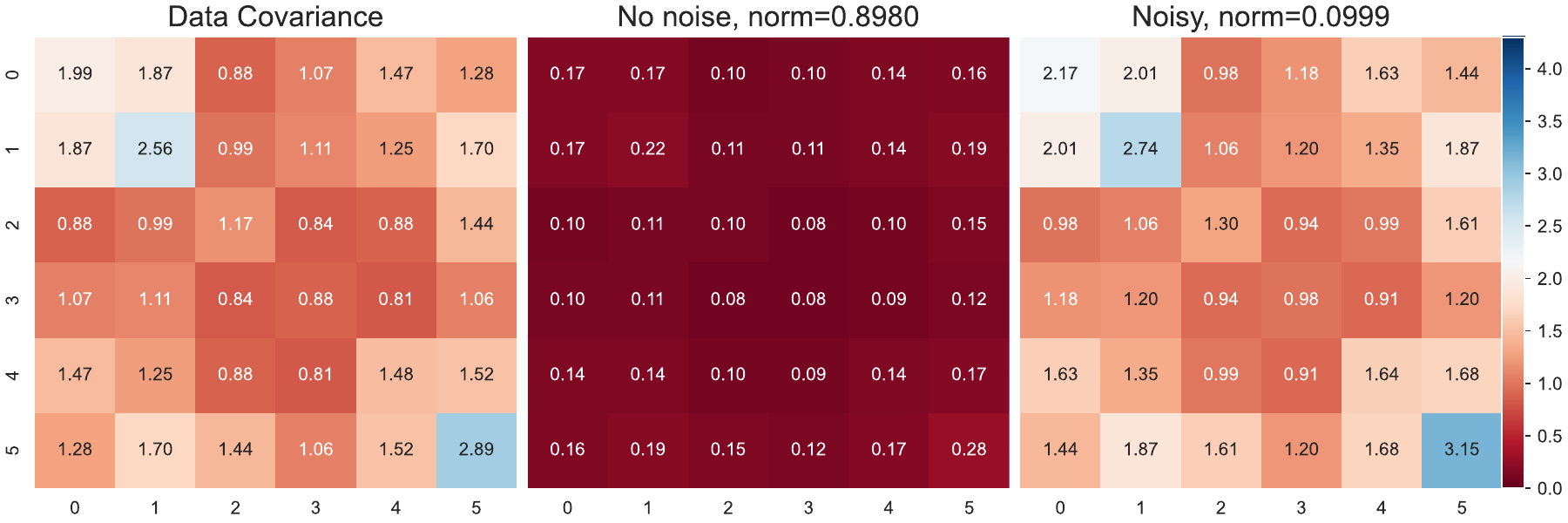}
    \caption{\textbf{Comparison of true data covariance and its estimates for the data generated from a Gaussian distribution.} Data covariance (\textit{left}), its estimates by gradient covariance matrices with clean targets (\textit{middle}), and with target noise of variance equal to batch size, $n=256$ (\textit{right}). We compare the results using the relative operator norm $r$ (Eq.~\ref{eq:norm_r}). }
    \label{fig:illustration_matrices}
\end{figure}

We first evaluate the proposed target noise injection on data drawn from a Gaussian distribution with a dense covariance matrix. Fig.~\ref{fig:illustration_matrices} compares the true data covariance with estimates from clean and noisy targets. While gradient covariance with clean targets partially captures the structure, it fails to reflect magnitude and complex dependencies. In contrast, adding noise to targets yields gradient covariances that closely match the true covariance (norm $r = 0.0999$). For this experiment, we used batch size $n=256$, noise variance equal to $n$, and $k=3125$ batches.

\begin{figure}[!t]
    \centering
    \includegraphics[width=\linewidth]{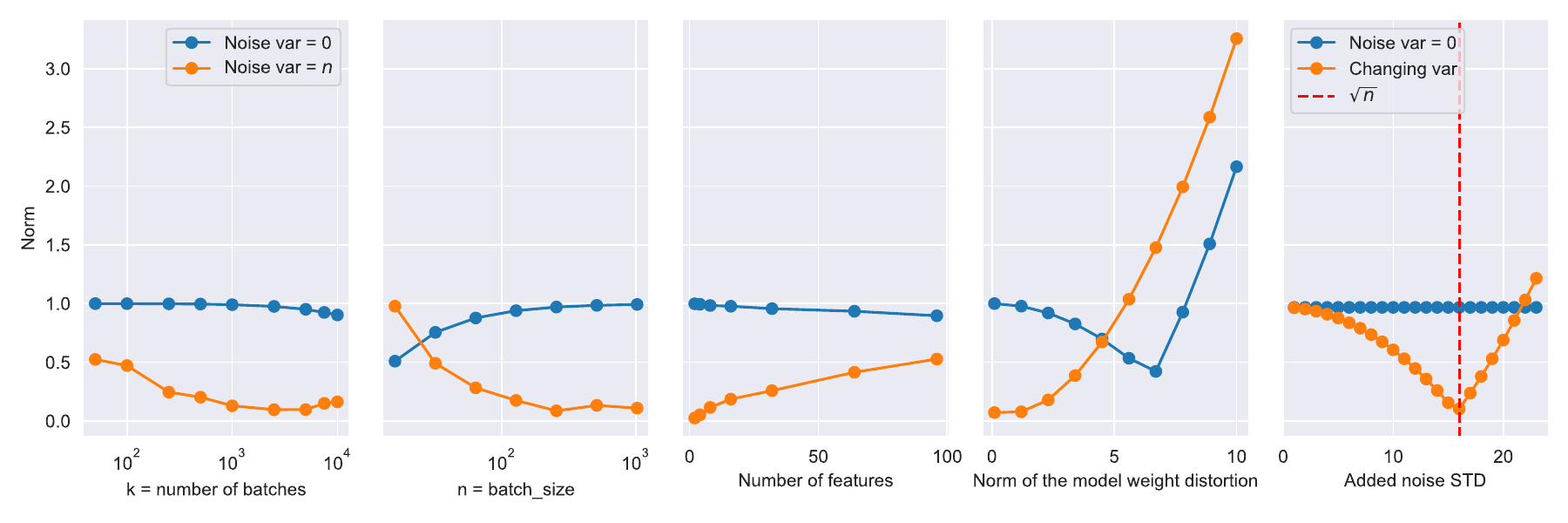}
    \caption{\textbf{Sensitivity of Hessian estimates.} Dependency of the estimates on (a) number of batches, (b) batch size, (c) number of features, (d) proximity of model weights to optimum, and (e) standard deviation of added target noise. We report norm $r$ (Eq.~\ref{eq:norm_r}). The estimate of the covariance matrix improves with a larger number of batches and larger batch sizes, and remains stable in the vicinity of the true model weights. The optimal added noise variance is equal to the batch size $n=256$. Here, the model is randomly initialised except in plot (d), the optimal weights are distorted by a random vector with norm $c$.}
    \label{fig:sensitivity_ext}
\end{figure}

We further investigated the sensitivity of the data covariance estimates to the number of available batches, batch size, the number of features in the input, and the standard deviation of the added target noise. For this experiment, we generated 800,000 samples from a Gaussian distribution with a diagonal covariance matrix and a standard deviation of 2. Fig.~\ref{fig:sensitivity_ext}a, b show that the estimates benefit from a larger number of batches $k$, given a sufficient batch size. We find that batch sizes above 64 work well; however, increasing the batch size $n$ is beneficial. We set $n=256$. As shown in Fig.~\ref{fig:sensitivity_ext}c, the dataset with a bigger number of input features would require more data to maintain the same quality of the estimates. The covariance estimate is robust to distortions in model parameters (Fig.~\ref{fig:sensitivity_ext}d). In this evaluation, the optimal values of model parameters are distorted by the addition of a random vector $\mathbf{v}$ with a norm equal to $c$, where $c$ changes from 0.1 to 10. Our covariance matrix estimate becomes more accurate as the parameter values approach the optimum. Lastly, Fig.~\ref{fig:sensitivity_ext}e shows that the optimal standard deviation (STD) for the added noise is equal to $\sqrt{n} = 16$. Further details can be found in Appendix~\ref{sec:data_gen}--~\ref{sec:batch_size_select}.

\subsection{Linear regression model}
\label{sec:exp_linear}

\begin{table*}[!tbh]
    \centering
    \small
    \caption{\textbf{Hessian estimation for public datasets with clean and noisy gradient covariance matrices.} Mean (and standard deviation) of test MSE and the relative operator norm $r$ are obtained with 10 random seeds.}
    \label{tab:lr_real}
    \vspace{2mm}
    \begin{tabular}{l|cc|cc|ccc|cc} %
        \multirow{2}{*}{\textbf{DATA}} & \multirow{2}{*}{ \textbf{FEAT.}} & \multirow{2}{*}{\textbf{SIZE}} & \multicolumn{2}{c}{\textbf{TEST MSE}} & \multicolumn{3}{|c}{\textbf{NORM} $r$, \textbf{linear}}  & \multicolumn{2}{|c}{\textbf{NORM} $r$, \textbf{MLP}} \\
         & & & linear & MLP & no noise & $\times n$ & noise = $n$  & no noise & noise = $n$ \\
        \hline %
        Wave & 48 & 288k & 1.7e-10 & 6.9e-8 & 1.0 (0.0) & 1.0 (6e-8) & \textbf{0.02} (0.01) & 1.0 (4e-8) & \textbf{0.02 (0.01)} \\
        Bike & 50 & 17k & 0.448 & 0.227 & 0.99 (4e-4) & 0.65 (0.02) & \textbf{0.29} (0.05) & 1.0 (2e-4) & \textbf{0.18 (0.04)} \\
        Housing & 8 & 20k & 0.322 & 0.286 & 0.99 (9e-4) & 2.05 (0.30) & \textbf{0.23} (0.24) & 0.99 (0.01) & \textbf{0.51 (0.35)} \\
        Wine & 10 & 5k & 0.474 & 0.421 & 0.99 (2e-3) & 0.42 (0.13) & \textbf{0.17} (0.10) & 0.99 (1e-3) & \textbf{0.22 (0.11)} \\
    \end{tabular}
\end{table*}

We further tested the estimate of the Hessian using noise injection to the targets on real-world datasets with linear regression models. We demonstrate the robustness of the proposed Hessian estimate during model training. In the experiment, the model parameters (weights) were updated after each batch with the noise-free gradients (gradients obtained from clean targets) using the Adam optimiser (\cite{kingma2014adam}) with a batch size of 64.

In this experiment, we utilised the following four publicly available datasets: wave energy converters (\cite{wave_energy_converters_534}), bike sharing (\cite{fanaee2014event}), California housing (\cite{pace1997sparse}), and wine quality (\cite{cortez2009modeling}).
To maintain consistency with theoretical sections, all input features were centred around zero by subtracting their mean. Similarly, we centred the targets; alternatively, a bias term can be included in the model to account for this. More details on data preprocessing and training can be found in Appendix~\ref{sec:data_preproc}. 

\begin{figure}[!tb]
    \centering
   \includegraphics[width=\linewidth]{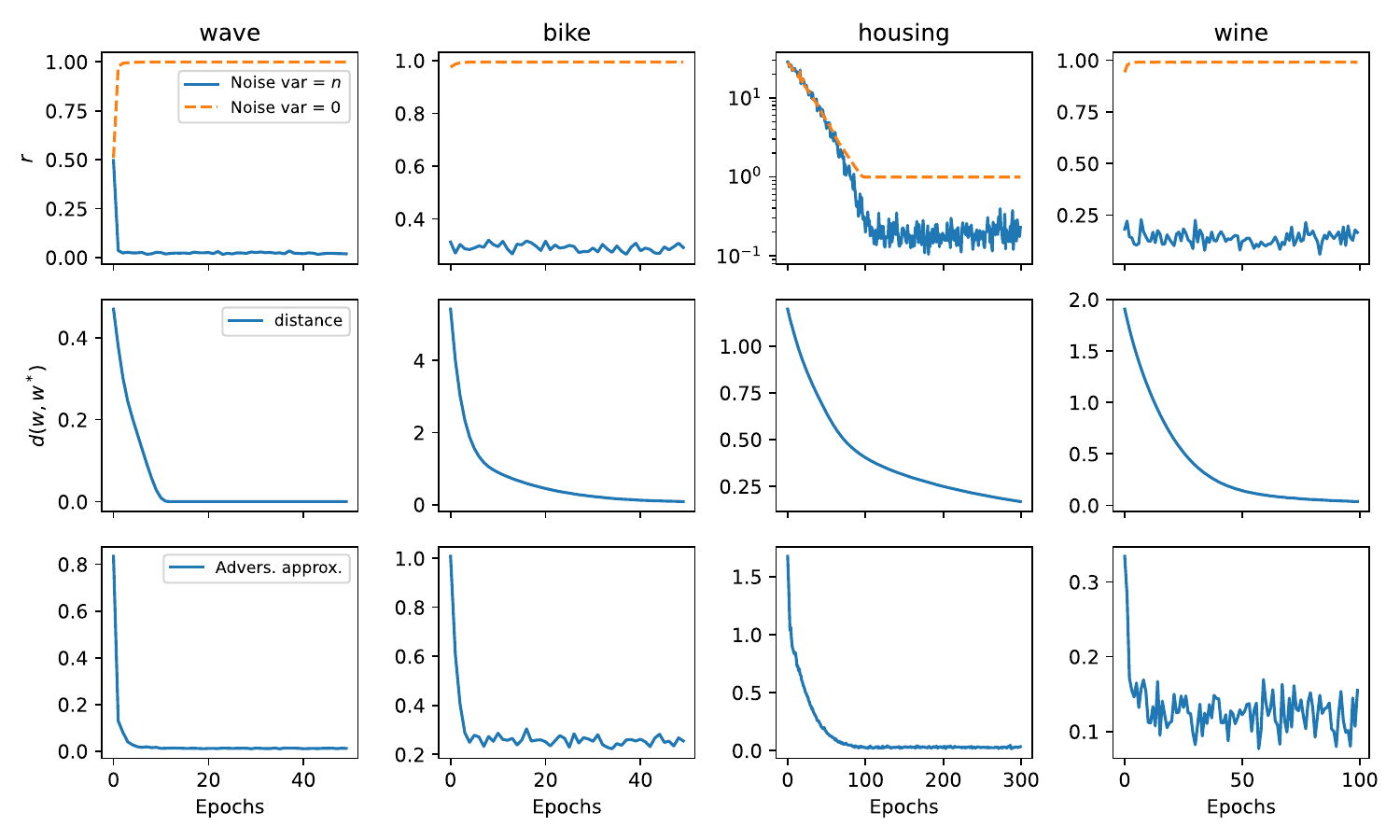}
    \caption{\textbf{Hessian estimate quality during model training.} \textit{Top}: the relative operator norm $r$ (Eq.~\ref{eq:norm_r} for the four studied datasets. \textit{Middle}: the corresponding distance between the analytical solution and the weight on the current epoch. \textit{Bottom}: Absolute error for the quadratic form approximation (Lemma~\ref{lemma:advrisk}) used in adversarial risk calculation (Lemma~\ref{lemma:advriskref}). The results correspond to the runs in Table~\ref{tab:lr_real} and are averaged across 10 random seeds. }
    \label{fig:norm_during_train}
\end{figure}

The characteristics of the tested datasets and the results by linear regression models are summarised in Table~\ref{tab:lr_real}. We reported test MSE for the trained models and the relative operator norm $r$ for three estimates of the Hessian: 1.~the gradient covariance of clean gradients (`no noise'); 2.~the estimate from 1. multiplied by the batch size $n$ (`$\times n$'); 3.~the proposed method for estimating the Hessian from the gradients with injected target noise with variance equal to $n$ (`noise = $n$'). 

The presented results demonstrate the efficacy of the proposed approach compared to the alternatives. Simple rescaling of the noise-free gradient covariance by multiplication (`$\times n$') yields an informative estimate, as seen for the bike and wine datasets; however, the estimate using the noisy gradient is always superior. In the experiment, the model weights are updated after evaluating each batch, introducing a slight mismatch between the mean gradients used for the covariance calculation. However, the estimate is robust to it, and these effects can be neglected in practice.

Furthermore, Fig.~\ref{fig:norm_during_train} shows that the proposed Hessian estimate is valid in a sufficiently large neighbourhood of the optimal solution; before the optimisation procedure fully converged (see the evolution of the norm $r$ during training, top row, and the corresponding distance between the current model weights and the analytical solution, middle row). This allows the utilisation of the obtained Hessian estimate as a preconditioner. Additionally, the estimated Hessian enables the quantification of adversarial risk by accurately approximating the quadratic form $\| w-w_0 \|_{\Sigma}^2$ as derived in Lemma~\ref{lemma:advrisk} (see Fig.~\ref{fig:norm_during_train}, bottom row).

\subsection{Extension to non-linear models}

\begin{figure}[!tb]
    \centering
    \includegraphics[width=\linewidth]{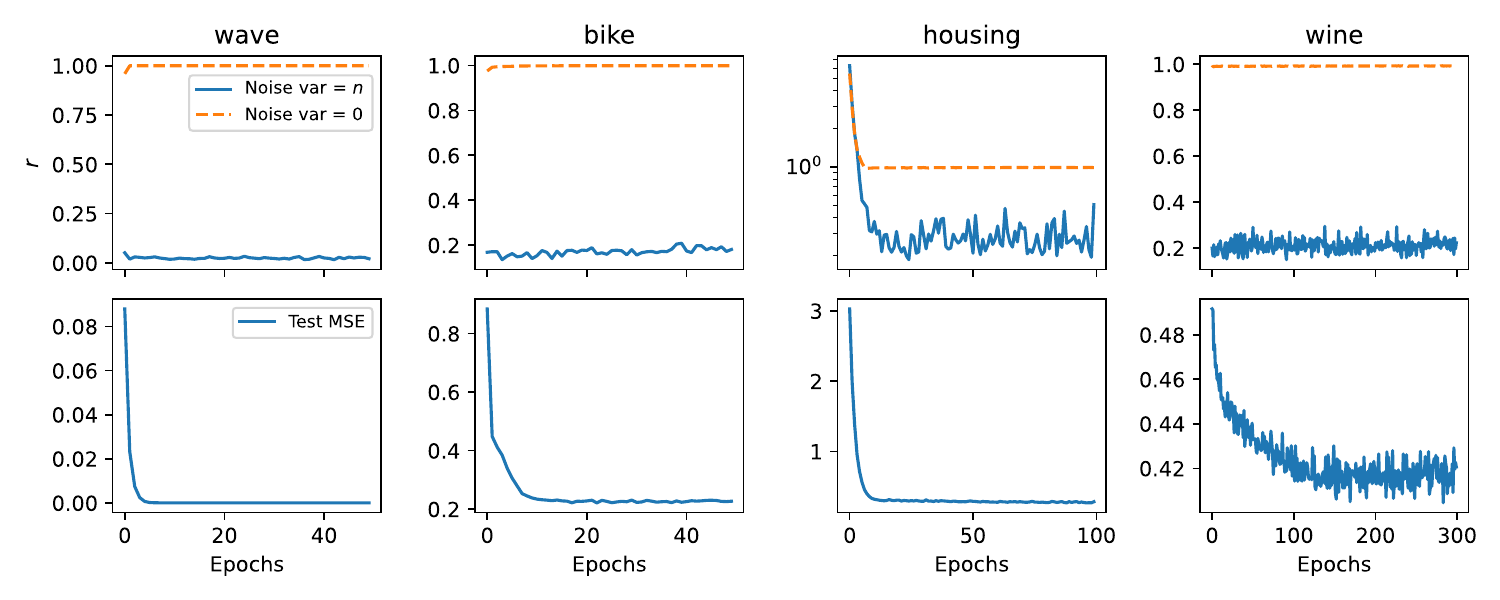}
    \caption{\textbf{Hessian estimate during MLP training with gradient covariance matrices.} \textit{Top}: the relative operator norm for the four studied datasets with noise-free and noisy gradients. \textit{Bottom}: Test MSE metric as the measure of convergence. The results correspond to the runs in Table~\ref{tab:lr_real} with an MLP. }
    \label{fig:norm_hessian_during_train}
\end{figure}

The results of the previous section can be extended to non-linear models for Hessian approximation. In the same experimental setup as in Section~\ref{sec:exp_linear}, we evaluated the Hessian estimates using a covariance matrix of clean and noisy gradients. Similar to the case of the linear model, the obtained Hessian estimates are stable in a neighbourhood of the optimal solution. They are more accurate compared to the covariance of noise-free gradients (see Table~\ref{tab:lr_real} and  Fig.~\ref{fig:norm_hessian_during_train}).
For all datasets, we trained an MLP with 64 hidden neurons and an output layer to avoid overfitting. Further details are provided in Appendix~\ref{sec:data_preproc}.

We also note that for an MLP with ReLU nonlinearities optimised with MSE loss, the blocks of the Hessian are target-independent (see Fig.~\ref{fig:h_diff} in Appendix~\ref{sec:hessian_relu_mlp}). Additionally, Appendix~\ref{sec:mlp_gen_diff_size_hessian} shows that noisy gradient covariance provides a reasonable estimate of the Hessian for MLPs of varying sizes on generated data.

\subsection{Preconditioning}

\begin{figure}[!t]
    \centering
    \includegraphics[width=\linewidth]{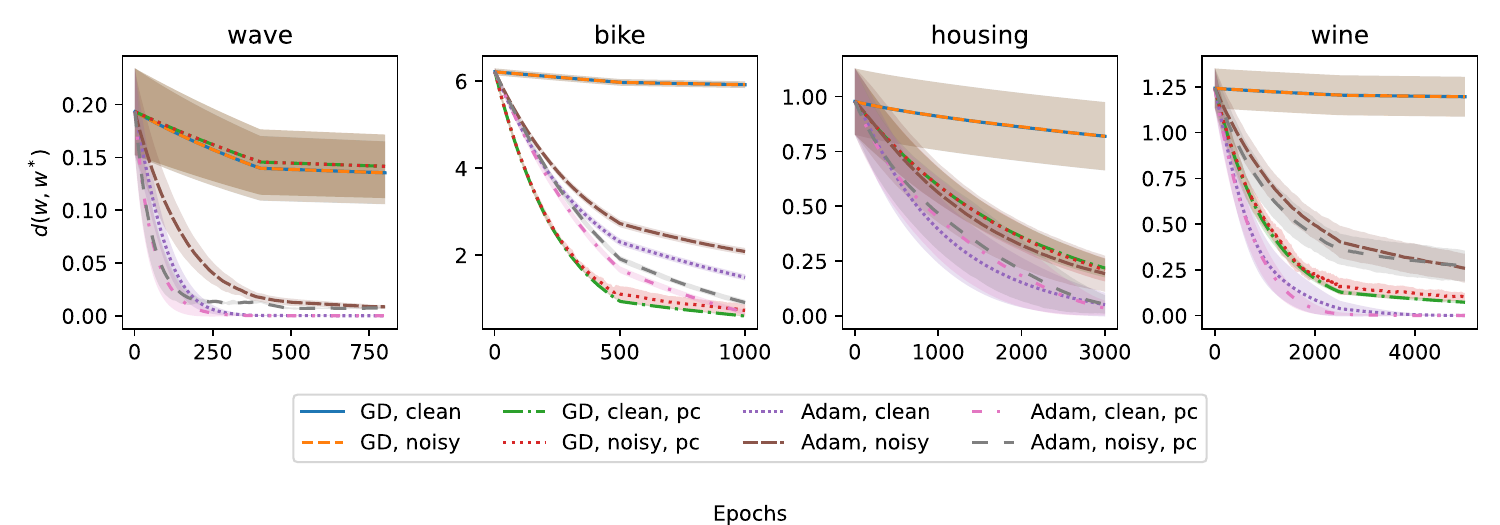}
    \caption{\textbf{Estimated Hessian as a preconditioner.} Optimisation with noisy mean gradient is close to the clean gradient (oracle). The inverse of the Hessian estimate as a covariance of noisy gradients is an effective preconditioner (pc) for both GD and Adam.}
    \label{fig:pc}
\end{figure}

In this experiment, we utilised noisy gradients both for estimating the Hessian and for parameter updates during training. We also compared it to the case when clean gradients are used for updates. We refer to this setting as "oracle" since two sets of targets (clean and noisy) may not be available.
We trained a linear model on datasets from Section~\ref{sec:exp_linear} with gradient descent (GD) and Adam as optimisers as described in Algorithm~\ref{alg:central}. Fig.~\ref{fig:pc} shows that 1.~the results for optimisation with noisy mean gradients are close to the ones with a clean mean gradient (oracle) and 2.~the estimated Hessian serves as an effective preconditioner for both optimisers, accelerating convergence in both clean and noisy gradient cases.
More results are in Appendix~\ref{sec:pc_results}.

\section{Conclusion and Future Work}
We showed that injecting Gaussian noise of variance $O(n)$ into the targets makes 
the empirical gradient covariance a reliable estimator of the Hessian 
$\Sigma$, with non-asymptotic operator-norm guarantees. The method preserves 
gradient accuracy and enables applications such as preconditioning, adversarial 
risk estimation, and gradient-only optimisation without access to raw data. We also showed extensions to non-linear models experimentally.  

For future work, we highlight two directions. First, while our experiments show that target noise improves Hessian estimation in neural networks, its theoretical properties remain open. Second, while adding noise with large variance helps recover the Hessian, it also increases the variance of the gradients, thereby requiring more data as the noise level grows. This raises the question of whether there exists a way to perform post hoc variance reduction to largely undo the additional variance.
% conjecture that the true batch size complexity for 
% uniform convergence within radius $R$ around $w_0$ should scale as 
% $n = \Omega(R^2/\epsilon)$, using only $k = \widetilde{\Omega}(d/\epsilon^2)$ batches. 
% The latter conjecture would imply that the batch size required by Theorem~\ref{thm:main} can be 
% reduced from $\widetilde{\Omega}(d^3/\epsilon^2)$ to $\widetilde{\Omega}(d/\epsilon)$, 
% suggesting that the present dependencies arise from our proof techniques rather than from the problem's intrinsic statistical complexity.

\subsubsection*{Acknowledgments} We thank the Research Council of Finland for funding, grants 364226 (Virtual Laboratory for Molecular Level Atmospheric Transformations (VILMA) Centre of Excellence) and 345704 (Finnish Center for Artificial Intelligence (FCAI)). We also thank the Helsinki Institute for Information Technology HIIT for funding. We thank the Finnish Computing Competence Infrastructure (FCCI) and CSC -- IT Center for Science, Finland -- for computational resources. The authors acknowledge the research environment provided by ELLIS Institute Finland.

\bibliography{ms}

\newpage
\section*{Checklist}

The checklist follows the references. For each question, choose your answer from the three possible options: Yes, No, Not Applicable.  You are encouraged to include a justification to your answer, either by referencing the appropriate section of your paper or providing a brief inline description (1-2 sentences). 
Please do not modify the questions.  Note that the Checklist section does not count towards the page limit. Not including the checklist in the first submission won't result in desk rejection, although in such case we will ask you to upload it during the author target period and include it in camera ready (if accepted).

\textbf{In your paper, please delete this instructions block and only keep the Checklist section heading above along with the questions/answers below.}

\begin{enumerate}

  \item For all models and algorithms presented, check if you include:
  \begin{enumerate}
    \item A clear description of the mathematical setting, assumptions, algorithm, and/or model. [\textbf{Yes] Throughout the text and in Appendix~\ref{app:notation}}
    \item An analysis of the properties and complexity (time, space, sample size) of any algorithm. \textbf{[Yes] Throughout the text and in Appendix}
    \item (Optional) Anonymized source code, with specification of all dependencies, including external libraries. \textbf{[Yes] Submitted as a zip supplement}
  \end{enumerate}

  \item For any theoretical claim, check if you include:
  \begin{enumerate}
    \item Statements of the full set of assumptions of all theoretical results. \textbf{[Yes]}
    \item Complete proofs of all theoretical results. \textbf{[Yes] 
    Proofs are provided in the Appendix~\ref{sec:proofs}.}
    \item Clear explanations of any assumptions. \textbf{[Yes] In the corresponding sections. } 
  \end{enumerate}

  \item For all figures and tables that present empirical results, check if you include:
  \begin{enumerate}
    \item The code, data, and instructions needed to reproduce the main experimental results (either in the supplemental material or as a URL). \textbf{[Yes] The code with the corresponding instructions is uploaded as a zip supplement.}
    \item All the training details (e.g., data splits, hyperparameters, how they were chosen). \textbf{[Yes] Details are provided in Appendix~\ref{sec:data_gen}~--~\ref{sec:data_preproc}.}
\item A clear definition of the specific measure or statistics and error bars (e.g., with respect to the random seed after running experiments multiple times). \textbf{[Yes] Section~\ref{sec:experiments}}
    \item A description of the computing infrastructure used. (e.g., type of GPUs, internal cluster, or cloud provider). \textbf{[Yes] We used an internal cluster with Xeon E5 2680, Xeon Gold 6148, and Xeon Gold 6248 CPUs (for the majority of the experiments) and NVIDIA V100 GPUs.}
  \end{enumerate}

  \item If you are using existing assets (e.g., code, data, models) or curating/releasing new assets, check if you include:
  \begin{enumerate}
    \item Citations of the creator If your work uses existing assets. \textbf{[Yes] The datasets are cited in Section~\ref{sec:experiments}.}
    \item The license information of the assets, if applicable. \textbf{[Yes] We used publicly available datasets in this work.}
    \item New assets either in the supplemental material or as a URL, if applicable. \textbf{[Yes] The code is submitted as a zip supplement.}
    \item Information about consent from data providers/curators. \textbf{[Not Applicable]}
    \item Discussion of sensible content if applicable, e.g., personally identifiable information or offensive content. \textbf{[Not Applicable]}
  \end{enumerate}

  \item If you used crowdsourcing or conducted research with human subjects, check if you include:
  \begin{enumerate}
    \item The full text of instructions given to participants and screenshots. \textbf{[Not Applicable]}
    \item Descriptions of potential participant risks, with links to Institutional Review Board (IRB) approvals if applicable. \textbf{[Not Applicable]}
    \item The estimated hourly wage paid to participants and the total amount spent on participant compensation. \textbf{[Not Applicable]}
  \end{enumerate}

\end{enumerate}

\onecolumn

\appendix

\section{PRELIMINARIES}

\subsection{Notation}\label{app:notation}
We consider a linear regression model where the response is generated as \( y = x^\top w_0 + \varepsilon \), with \( x \in \mathbb{R}^d \) denoting the input vector, \( w_0 \in \mathbb{R}^d \) the true parameter, and \( \varepsilon \sim \mathcal{N}(0, \sigma^2) \) independent noise. We assume \( \mathbb{E}[x] = 0 \) and denote the population covariance by \( \Sigma = \mathbb{E}[x x^\top] \), which is assumed to be positive definite. While our theoretical results hold under sub-Gaussian assumptions, we also consider the special case where \( x \sim \mathcal{N}(0, \Sigma) \).

The dataset is partitioned into \( k \) batches, each containing \( n \) i.i.d.\ samples. For batch \( j \in [k] \), the design matrix is \( X^{(j)} \in \mathbb{R}^{n \times d} \) and the response vector is \( y^{(j)} \in \mathbb{R}^n \). The empirical least-squares loss is \( L^{(j)}(w) = \frac{1}{2n} \|X^{(j)} w - y^{(j)}\|^2 \), with gradient \( \nabla L^{(j)}(w) \), Hessian $\nabla^2 L^{(j)}(w)$, and batch minimizer \( w^{(j)}_* = \arg\min_w L^{(j)}(w) \). The empirical gradient covariance is
\[
S_g(w) = \frac{1}{k} \sum_{j=1}^k \left( \nabla L^{(j)}(w) - \frac{1}{k} \sum_{\ell=1}^k \nabla L^{(\ell)}(w) \right) \left( \nabla L^{(j)}(w) - \frac{1}{k} \sum_{\ell=1}^k \nabla L^{(\ell)}(w) \right)^\top.
\]
The empirical covariance for batch $j$ is \( H^{(j)} = \frac{1}{n} (X^{(j)})^\top X^{(j)} \), and the population covariance is \( \Sigma \). 

% We define the empirical covariance by $\widehat{\Sigma} = \frac{1}{nk}\sum_{j=1}^k \sum_{i=1}^n x_i^{(j)}(x_i^{(j)})^\top$.

For any vector \( x \in \mathbb{R}^d \), we denote by \( x_{[i]} \in \mathbb{R} \) the \( i \)th coordinate of \( x \), for every \( i \in [d] \). Similarly, for any matrix \( A \in \mathbb{R}^{d \times d} \), we denote by \( A_{[i,j]} \in \mathbb{R} \) the entry in the \( i \)th row and \( j \)th column of \( A \), for every \( i, j \in [d] \).

We use \( \|\cdot\|_2 \) for the Euclidean norm, \( \|\cdot\|_{\mathrm{op}} \) for the operator norm (largest singular value), and \( \|\cdot\|_F \) for the Frobenius norm. For symmetric matrices, \( \lambda_{\min}(\cdot) \) and \( \lambda_{\max}(\cdot) \) denote the smallest and largest eigenvalues. We write \( \mathbb{E}[\cdot] \) and \( \mathbb{P}[\cdot] \) for expectation and probability, respectively.

We denote by $\mathrm{SG}(\sigma^2)$ the class of sub-Gaussian random variables 
with parameter $\sigma^2$, and by $\mathrm{SG}_d(\sigma^2)$ the class of sub-Gaussian 
random vectors in $\mathbb{R}^d$ with parameter $\sigma^2$. Similarly, we denote by 
$\mathrm{SE}(\nu^2, \alpha)$ the class of sub-exponential random variables with 
parameters $(\nu^2, \alpha)$, and by $\mathrm{SE}_d(\nu^2, \alpha)$ the class of 
sub-exponential random vectors in $\mathbb{R}^d$ with parameters $(\nu^2, \alpha)$. Constants such as \( \epsilon \in (0,1) \), \( \delta \in (0,1) \), and \( c > 0 \) appear in concentration inequalities and sample complexity results.

We use $\mathcal{O}(\cdot)$ and $\Omega(\cdot)$ to denote upper and lower bounds 
up to absolute constants. The notations $\widetilde{\mathcal{O}}(\cdot)$ and 
$\widetilde{\Omega}(\cdot)$ are used similarly, but with logarithmic factors hidden. The symbol \( \gtrsim \) denotes informal lower bounds up to absolute constants, i.e., \( a \gtrsim b \) means \( a \geq C b \) for some constant \( C > 0 \); the reverse relation \( a \lesssim b \) denotes \( a \leq C b \).

\subsection{Useful Prior Results}
Unless stated otherwise, most of the results in this section are taken from \cite{vershynin2018high} and \cite{wainwright2019high}.

\begin{definition}
    [$\varepsilon$-net] Let $(T,d)$ be a metric space. Consider a set $K \subset T$ and a number $\varepsilon > 0$. A subset $N_\epsilon \subset K$ is called an $\varepsilon$-net of $K$ if every point in $K$ is within distance $\varepsilon$ of some point of $N_\epsilon$, i.e.
\begin{align}
\forall x \in K \;\exists x_0 \in N_\epsilon : d(x,x_0) \le \varepsilon.
\end{align}
\end{definition}

\begin{definition}
    [Covering number] The smallest cardinality of an $\varepsilon$-net of $K$ is called the \textit{covering number} of $K$ and is denoted $\mathcal{N}(K,\varepsilon)$.
\end{definition}

\begin{definition}
    [Sub-Gaussian random vector]
    A vector $\mathbf{x} \in \mathbb{R}^d$ with $\mathbb{E}[x] = \mu$ is vector sub-Gaussian 
($\mathbf{x} \in SG_d(\sigma^2)$) with parameter $\sigma$ if for all 
$v \in \mathbb{R}^d$ such  that $||\mathbf{v}|| = 1$ we have

\begin{align}
\mathbb{E}\left[\exp\left(\mathbf{v}^{\top}(\mathbf{x} - \mu)\right)\right] 
\leq \exp\left(\lambda^2 \sigma^2 / 2\right)
\end{align}

\end{definition}

\begin{definition}
    [Sub-Exponential random variable] \textit{Centered random variable $X \in SE(\nu^2, \alpha)$ with parameters $\nu, \alpha > 0$ if:}
\begin{align}
    \mathbb{E} \left[\exp(\lambda X)\right] \leq \exp(\lambda^2 \nu^2/2), \forall \lambda : |\lambda| < \frac{1}{\alpha} .
\end{align}
\end{definition}

\begin{definition} [Sub-Exponential random vector]
     A vector $\mathbf{x} \in \mathbb{R}^d$ with $\mathbb{E}[x] = \mu$ is vector sub-Exponential 
($\mathbf{x} \in SE_d(\nu^2, \alpha)$) with parameter $\nu^2, \alpha > 0$ if for all 
$v \in \mathbb{R}^d$ such  that $||\mathbf{v}|| = 1$ we have
\begin{align}
    \mathbb{E} \left[\exp(\lambda (x - \mu)\right] \leq \exp(\lambda^2 \nu^2/2), \forall \lambda : |\lambda| < \frac{1}{\alpha} .
\end{align}
\end{definition}

\begin{theorem}[Sub-Gaussian Squared]\label{subgaussian-squared}\cite{honorio2014tight}
    For a sub-gaussian random variable $X \in SG(\sigma^2)$ with $\EX[X] = 0$, $X^2 - \EX[X^2]$ is sub-exponential, i.e., $X^2 - \EX[X] \in SE(32\sigma^4,4\sigma^2)$.
\end{theorem}

\begin{theorem}[Tail bound for Sub-Exponential Random Variables]\label{subexp-concentrate}
     Let $X \in SE(\nu^2, \alpha)$. Then:
\begin{align}
    \mathbb{P}(|X-\mu| \geq t) \leq 
    \begin{cases}
        2e^{-t^2 / (2\nu^2)}, & 0 < t \leq \frac{\nu^2}{\alpha} \quad \text{(Sub-Gaussian behavior)} \\
        2e^{-t/(2\alpha)}, & t > \frac{\nu^2}{\alpha}
    \end{cases}
\end{align}

\textit{It can be equivalently stated as:}
\begin{align}
    \mathbb{P}(|X-\mu| \geq t) \leq e^{-\frac{1}{2}\min\left\{\frac{t^2}{\nu^2}, \frac{t}{\alpha}\right\}}
\end{align}
\end{theorem}

\begin{theorem}[Composition property of Sub-Exponential random variables]\label{subexp-compos}
    Let $X_1, \dots, X_n$ be independent random variables such that $\mathbb{E}X_i = \mu_i$ and $X_i \in SE(\nu_i^2, \alpha_i)$. Then
\begin{align*}
    \sum_{i=1}^n (X_i - \mu_i) \in SE\left(\sum_{i=1}^n \nu_i^2, \max_i \alpha_i\right)
\end{align*}
\end{theorem}

\begin{lemma}[Gaussian is sub-Gaussian]\label{gaussian_is_sub}
    If $X \sim \mathcal{N}(0, \Sigma)$, then $X$ is a sub-Gaussian random vector with parameter $\|\Sigma\|_{\text{op}}$ ($X \in SG_d(\|\Sigma\|_\mathrm{op})$.
\end{lemma}

\begin{theorem}[Covering numbers of the Euclidean ball]\label{covering-number}
     The covering numbers of the unit Euclidean ball $B_2^d$ satisfy the following for any $\varepsilon > 0$:
\begin{align}
\left(\frac{1}{\varepsilon}\right)^d \le \mathcal{N}(B_2^d, \varepsilon) \le \left(\frac{2}{\varepsilon}+1\right)^d. \tag{4.17}
\end{align}
The same upper bound is true for the unit Euclidean sphere $S^{d-1}$.
\end{theorem}

\begin{theorem}[Gaussian norm Concentration]\label{gaussdist} \cite{speicher2023high} 
Consider a \( d \)-dimensional standard Gaussian random vector 
\( x \sim \mathcal{N}(0, I_d) \). Then, for \( 0 \leq \varepsilon \leq \sqrt{d} \),

\begin{align}
\mathbb{P} \left\{\left| \|x\| - \sqrt{d} \right| \geq \varepsilon \right\}
\leq 2 \exp\left( -\frac{\varepsilon^2}{16} \right)
\end{align}
    
\end{theorem}

\begin{theorem}[Weyl's Theorem: Spectral Stability]\label{weyl}
    For any  symmetric matrices $A,B \in \mathbb{R}^{d \times d}$ we have:
    \begin{align}
\bigl|\lambda_i(B)-\lambda_i(A)\bigr| \;\le\; \|A-B\|_\mathrm{op},
\qquad i=1,\dots,d.
\end{align}
\end{theorem}

\begin{lemma}[Gaussian Sample Covariance Concentration]\label{gausscov}
    Let \(\mathbf{x}_i \overset{\text{i.i.d.}}{\sim} \mathcal{N}(0, \Sigma)\) for \(i = 1, \dots, n\), where \(\mathbf{x}_i \in \mathbb{R}^d\). Defining the empirical covariance matrix 
\(\hat{\Sigma} \triangleq \frac{1}{n} \sum_{i=1}^n \mathbf{x}_i \mathbf{x}_i^\top\), as long as 
\(n \geq \frac{20}{\epsilon^2}(d + \log(1/\delta))\), we have with probability at least \(1 - \delta\),
\begin{align}
\frac{\|\hat{\Sigma} - \Sigma\|_{\text{op}}}{\|\Sigma\|_{\text{op}}} \leq \epsilon
\end{align}

\end{lemma}

\begin{theorem}[Sub-Gaussian Sample Covariance Concentration]\label{subgausscov}
    Let $\mathbf{x}_1, \dots, \mathbf{x}_n$ be an i.i.d. sequence of $\sigma^2$ sub-Gaussian random vectors such that $\mathrm{Cov}[\mathbf{x}_1] = \Sigma$, and let 
$\hat{\Sigma} := \frac{1}{n} \sum_{i=1}^n \mathbf{x}_i \mathbf{x}_i^\top$ be the empirical covariance matrix. Then there exists a universal constant 
$C_1 > 0$ such that, for $\delta \in (0,1)$, with probability at least $1 - \delta$,

\begin{align}
\frac{\|\hat{\Sigma} - \Sigma\|_{\text{op}}}{\sigma^2}
\leq C_1 \max \left\{
\sqrt{\frac{d + \log(2/\delta)}{n}},
\frac{d + \log(2/\delta)}{n}
\right\}.
\end{align}
Therefore, for $\epsilon < 1$, there exists a universal constant $C > 0$, such that as long as we have $n \geq C \frac{\sigma^2}{\epsilon^2} \left( d + \log\left( 1/\delta \right) \right)$, then
\begin{align}
    \|\hat{\Sigma} - \Sigma\|_{\text{op}}/\sigma^2 \leq \epsilon.
\end{align}
\end{theorem}

\begin{theorem}[Isserlis's theorem]\label{Isserlis}\cite{isserlis1918formula}
Let $(X_1, \ldots, X_n)$ be a zero-mean multivariate normal random vector. Then for even \( n \), the expectation of the product is given by
\[
\mathbb{E}[X_1 X_2 \cdots X_n] = \sum_{p \in P_n^2} \prod_{\{i,j\} \in p} \mathbb{E}[X_i X_j] = \sum_{p \in P_n^2} \prod_{\{i,j\} \in p} \operatorname{Cov}(X_i, X_j),
\]
where the sum is over all pairings \( p \) of the set \( \{1, \ldots, n\} \), i.e., all distinct ways of partitioning \( \{1, \ldots, n\} \) into unordered pairs \( \{i,j\} \), and the product is over the pairs in \( p \).
\end{theorem}

\section{PROOFS}
\label{sec:proofs}

\subsection{Proof of Example~\ref{thm:equalhessian}}
\begin{proof}
Using Taylor's theorem, each loss function can be expressed as
\[
L^{(i)}(w) = L^{(i)}(w_*^{(i)}) + \frac{1}{2} (w - w_*^{(i)})^{\top} \Sigma (w - w_*^{(i)}),
\]
since $\nabla L^{(i)}(w_*^{(i)}) = 0$ and $\nabla^2 L^{(i)} = \Sigma$. Differentiating, we find
\[
\nabla L^{(i)}(w) = \Sigma (w - w_*^{(i)}).
\]
Averaging over $i$ gives
\[
\frac{1}{k} \sum_{i=1}^k \nabla L^{(i)}(w) = \Sigma \left(w - \widehat{w}\right).
\]
The gradient covariance matrix is defined as
\begin{align*}
S_g(w) &= \frac{1}{k} \sum_{i=1}^k \left(\nabla L^{(i)}(w) - \frac{1}{k} \sum_{j=1}^k \nabla L^{(j)}(w)\right)\left(\nabla L^{(i)}(w) - \frac{1}{k} \sum_{j=1}^k \nabla L^{(j)}(w)\right)^{\top} \\
&= \frac{1}{k} \sum_{i=1}^k \left( \Sigma(w - w_*^{(i)}) - \Sigma(w - \widehat{w}) \right) \left( \Sigma(w - w_*^{(i)}) - \Sigma(w - \widehat{w}) \right)^{\top} \\
&= \frac{1}{k} \sum_{i=1}^k \left( \Sigma(\widehat{w} - w_*^{(i)}) \right) \left( \Sigma(\widehat{w} - w_*^{(i)}) \right)^{\top} \\
&= \Sigma \left( \frac{1}{k} \sum_{i=1}^k (w_*^{(i)} - \widehat{w})(w_*^{(i)} - \widehat{w})^{\top} \right) \Sigma \\
&= \Sigma \widehat{S}_{w_*} \Sigma.
\end{align*}
\end{proof}

\subsection{Proof of Theorem~\ref{thm:main}}

\begin{proof*}
First, note that we have $k$ batches, each containing $n$ data points. We will derive the values of $k$ and $n$ later. 

For each batch $j$ we define $H^{(j)} = \frac{1}{n}\sum_{i=1}^nx_i^{(j)} (x_i^{(j)})^{\top}$.
Note that based on Theorem~\ref{subgausscov} for $n \geq \frac{C d^2}{c^4 \epsilon^2}(d+\log(k/\delta))$ (where $C$ is an absolute constant), for every batch we have with probability $\geq 1-\frac{\delta}{k}$ that $||H^{(j)} - \Sigma||_\mathrm{op} \leq \frac{c^2 \epsilon}{d}$. Hence, by union bounding over the $k$ batches, with probability at least $1 - \delta$ we have
\begin{align}
    ||H^{(j)} - \Sigma||_\mathrm{op} \leq \frac{c^2 \epsilon}{d} ~~~ \text{for all}~ j\in [k].
\end{align}
We call this event $\mathcal{E}_1$ and through the rest of the proof, we are in the regime that $\mathcal{E}_1$ has occurred.

For simplicity we define $E^{(j)} := H^{(j)} - \Sigma$ and $\widehat{w} := \frac{1}{k}\sum_{i=1}^k w^{(i)}_*$. Now we write
\begin{align}
    S_g(w) &= \frac{1}{k}\sum_{i=1}^k(\nabla L^{(i)}(w) - \frac{1}{k}\sum_{j=1}^k\nabla L^{(j)}(w)) (\nabla L^{(i)}(w) - \frac{1}{k}\sum_{j=1}^k\nabla L^{(j)}(w))^{\top}
    \\
    &= \frac{1}{k}\sum_{i=1}^k \left[H^{(i)}(w - w_*^{(i)}) - \frac{1}{k}\sum_{j=1}^kH^{(j)}(w - w_*^{(j)})\right] 
    \\
     &~~~~~~~~~~~\left[H^{(i)}(w - w_*^{(i)}) - \frac{1}{k}\sum_{j=1}^kH^{(j)}(w - w_*^{(j)})\right]^{\top} 
    \\
    &= \frac{1}{k}\sum_{i=1}^k\left[(\Sigma + E^{(i)})(w - w_*^{(i)}) - \frac{1}{k}\sum_{j=1}^k(\Sigma + E^{(j)})(w - w_*^{(j)})\right]
    \\
    &~~~~~~~~~~~\left[(\Sigma + E^{(i)})(w - w_*^{(i)}) - \frac{1}{k}\sum_{j=1}^k(\Sigma + E^{(j)})(w - w_*^{(j)})\right]^{\top}
    \\
    &= \frac{1}{k}\sum_{i=1}^k \left[\Sigma(\widehat{w} - w^{(i)}_*) + E^{(i)}(w-w_*^{(i)}) - \frac{1}{k}\sum_{j=1}^k  E^{(j)}(w - w_*^{(j)})\right]
    \\
    &~~~~~~~~~~~\left[\Sigma(\widehat{w} - w^{(i)}_*) + E^{(i)}(w-w_*^{(i)}) - \frac{1}{k}\sum_{j=1}^k  E^{(j)}(w - w_*^{(j)})\right]^{\top}
    \\
    &= \frac{1}{k}\sum_{i=1}^k \left[\Sigma(\widehat{w} -w_0 + w_0 - w^{(i)}_*) + E^{(i)}(w-w_*^{(i)}) - \frac{1}{k}\sum_{j=1}^k  E^{(j)}(w - w_*^{(j)})\right]
    \\
    &~~~~~~~~~~~\left[\Sigma(\widehat{w} -w_0 + w_0 - w^{(i)}_*) + E^{(i)}(w-w_*^{(i)}) - \frac{1}{k}\sum_{j=1}^k  E^{(j)}(w - w_*^{(j)})\right]^{\top}
\end{align}
Hence, we can write:
\begin{align}
    S_g(w) &= \frac{1}{k}\sum_{i=1}^k \left[\Sigma(\widehat{w} -w_0) + \Sigma(w_0 - w^{(i)}_*) + E^{(i)}(w-w_*^{(i)}) - \frac{1}{k}\sum_{j=1}^k  E^{(j)}(w - w_*^{(j)})\right]
    \\
    &~~~~~~~~~~~\left[\Sigma(\widehat{w} -w_0) + \Sigma(w_0 - w^{(i)}_*) + E^{(i)}(w-w_*^{(i)}) - \frac{1}{k}\sum_{j=1}^k  E^{(j)}(w - w_*^{(j)})\right]^{\top}
\end{align}
By expanding, we have 16 terms:
\begin{align}
        S_g(w) &= \underbrace{\Sigma(\widehat{w} - w_0)(\widehat{w} - w_0)^{\top} \Sigma}_A + \underbrace{\frac{1}{k}\sum_{i=1}^k \Sigma (w^{(i)}_* - w_0)(w^{(i)}_* - w_0)^{\top}}_B \Sigma 
    \\
    &+ \underbrace{\frac{1}{k}\sum_{i=1}^k \Sigma(\widehat{w}-w_0)(w_0-w_*^{(i)})^{\top}\Sigma}_C + 
    \underbrace{\frac{1}{k}\sum_{i=1}^k \Sigma(w_0-w_*^{(i)})(\widehat{w}-w_0)^{\top}\Sigma}_D +
    ...
\end{align}

 Before analysing the terms, we need some auxiliary lemmas. 
 \end{proof*}

\begin{auxproofbox}
\begin{lemma}\label{spectrallemma}
    Under event $\mathcal{E}_1$, for all batches $j \in [k]$, we have
    \begin{align}
        ||(H^{(j)})^{-1} - \Sigma^{-1}|| \leq \frac{2\epsilon}{d}.
    \end{align}
\end{lemma}

 \begin{proof*}
 Note that for all $j \in [k]$ we have $(H^{(j)})^{-1}$ is close to $\Sigma$. Indeed for any $j \in [k]$ we have:
\begin{align}
    (H^{(j)})^{-1} - \Sigma^{-1} = \Sigma^{-1}(\Sigma-H^{(j)})(H^{(j)})^{-1},
\end{align}
hence we have 
\begin{align}
    ||(H^{(j)})^{-1} - \Sigma^{-1}||_\mathrm{op} &\leq ||\Sigma^{-1}||_\mathrm{op}||\Sigma-H^{(j)}||_\mathrm{op}||(H^{(j)})^{-1}||_\mathrm{op} 
    \\
    &\leq \frac{1}{c} \frac{c^2\epsilon}{d}||(H^{(j)})^{-1}||_\mathrm{op}.
\end{align}
Now note that as $||H^{(j)} - \Sigma||_\mathrm{op} \leq \frac{c^2\epsilon}{d}$, by Weyl's theorem (Theorem~\ref{weyl}) we have 
\begin{align}
    |\lambda_d(H^{(j)}) - \lambda_d(\Sigma)| \leq \frac{c^2\epsilon}{d}.
\end{align}
Hence, we have $||(H^{(j)})^{-1}||_\mathrm{op} \leq \frac{1}{\lambda_d(H^{(j)}) - (c^2\epsilon/d)} = \frac{1}{c - (c^2\epsilon/d)} \leq 2/c$. Then we can conclude:
\begin{align}
    ||(H^{(j)})^{-1} - \Sigma^{-1}|| \leq \frac{1}{c} \frac{c^2\epsilon}{d}\frac{2}{c} = \frac{2\epsilon}{d}.
\end{align}
\end{proof*}
\end{auxproofbox}

\begin{auxproofbox}
\begin{lemma}\label{closeto0}
    Under event $\mathcal{E}_1$, with probability at least $1 - \delta$, event $\mathcal{E}_2$ as defined by 
    \begin{align}
        ||w_*^{(i)} - w_0||_2 \leq\ \sqrt{8d/c}~~\text{for all}~~i \in [k].
    \end{align}
    happens, if we have
    \begin{align}
        d \geq 64[\log(2k) + \log(1/\delta)].
    \end{align}
\end{lemma}

\begin{proof*}
    
We need to show that with high probability the distances between all $w_*^{(i)}$s and $w_0$ are bounded (i.e., they are on the order of $\approx \sqrt{d}$). First note that
\begin{align}
    w_*^{(i)} - w_0 \sim \mathcal{N}(0,(H^{(i)})^{-1}) \Leftrightarrow 
    w_*^{(i)} - w_0 = (H^{(i)})^{-\frac{1}{2}}Z^{(i)} ~~\text{s.t.} ~~ Z^{(i)} \sim \mathcal{N}(0,I).
\end{align}
Hence, we can write:
\begin{align}
     ||w_*^{(i)} - w_0||_2 \leq ||(H^{(i)})^{-\frac{1}{2}}||_\mathrm{op} ||Z^{(i)}||_2 \leq \sqrt{2/c} ||Z^{(i)}||_2.
\end{align}
By Theorem~\ref{gaussdist} we have $\mathbb{P} \left\{\left| \|Z^{(i)}\| - \sqrt{d} \right| \geq \sqrt{d}/2 \right\}
\leq 2 \exp\left( -\frac{d}{64} \right)$, hence by union bound over all batches, we have
\begin{align}
    \mathbb{P} \left\{\exists i\in [k]:\left| \|Z^{(i)}\| - \sqrt{d} \right| \geq \sqrt{d}/2 \right\}
\leq 2k \exp\left( -\frac{d}{64} \right).
\end{align}
Now, if we consider the failure probability for this event as $\delta > 0$, we have:
\begin{align}
    2k \exp\left( -\frac{d}{64} \right) \leq \delta
    \Leftrightarrow  d \geq 64[\log(2k) + \log(1/\delta)]
\end{align}

So the final result is that, with probability at least $1-\delta$ we have 
\begin{align}
    ||w_*^{(i)} - w_0||_2 \leq \sqrt{2/c}\frac{3}{2}\sqrt{d} \leq\ \sqrt{8d/c}~~\text{for all}~~i \in [k].
\end{align}
\end{proof*}
\end{auxproofbox}

Another Lemma that we need for the rest of the proof is that under the event that $\mathcal{E}_1$ is true, $||\widehat{w}-w_0||_2$ is small.

\begin{auxproofbox}
\begin{lemma}\label{hatsmall}
    Under event $\mathcal{E}_1$, if $k \geq d/\epsilon^2$, with probability at least $1 - \delta$, event $\mathcal{E}_3$ as defined by 
    \begin{align}
    ||(\frac{1}{k}\sum_{j=1}^k w_*^{(i)}) - w_0||_2 \leq 
    \frac{2\epsilon}{\sqrt{c}}.
\end{align}
happens.
\end{lemma}

\begin{proof*}
Note that for all $i \in [k]$ we have $w_*^{(i)} \sim \mathcal{N}(w_0, {(H^{(i)})}^{-1})$. Hence because of independence of $w_*^{(i)}$s  we have 
\begin{align}
    (\frac{1}{k}\sum_{j=1}^k w_*^{(i)}) - w_0 \sim \mathcal{N}(0, \frac{1}{k^2}\sum_{i=1}^k {(H^{(i)})}^{-1}).
\end{align}
By the triangle inequality, we have
\begin{align}
    ||\frac{1}{k^2}\sum_{i=1}^k {(H^{(i)})}^{-1}||_\mathrm{op} \leq 2/ck.
\end{align}
Hence, using a similar argument as the previous lemma, by defining $Z \sim \mathcal{N}(0,I)$, we have
\begin{align}
     (\frac{1}{k}\sum_{j=1}^k w_*^{(i)}) - w_0 = (\frac{1}{k}\sum_{i=1}^k {(H^{(i)})}^{-1})^{1/2}Z 
\end{align}
Now similar to proof of Lemma~\ref{closeto0} we can prove that for large enough $d$, with probability at least $1-\delta$ we have 
\begin{align}
    ||(\frac{1}{k}\sum_{j=1}^k w_*^{(i)}) - w_0||_2 \leq \sqrt{\frac{4d}{ck}}.
\end{align}
And because we will set $k \geq d/\epsilon^2$, we have 
\begin{align}
    ||(\frac{1}{k}\sum_{j=1}^k w_*^{(i)}) - w_0||_2 \leq 
    \frac{2\epsilon}{\sqrt{c}}.
\end{align}
\end{proof*}
\end{auxproofbox}

\begin{auxproofbox}
\begin{lemma}\label{wclose}
    Under event $\mathcal{E}_1$ and $\mathcal{E}_2$, for all $i \in [k]$ we have
    \begin{align}
        \|w_*^{(i)}-w_0\|_2 \leq3\sqrt{d}
    \end{align}
\end{lemma}
\begin{proof*}
By triangle inequality it's easy to see that for all $i$ we have $||w-w_*^{(i)}||_2 \leq 4\sqrt{d}$ as by assumption we have $||w-w_0||_2 \leq \sqrt{d}$ and we proved in Lemma~\ref{closeto0} that for all $i \in [k]$ we have $\|w_*^{(i)}-w_0\|_2 \leq3\sqrt{d}$.
\end{proof*}
\end{auxproofbox}

\begin{continuedproof*}{Theorem~\ref{thm:main}}

We now continue the proof. We condition on happening of events $\mathcal{E}_1$, $\mathcal{E}_2$, and $\mathcal{E}_3$. 

We derive each term in order to prove $B \approx \Sigma$ and for all other terms $X$ we have $||X||_\mathrm{op} \lesssim \epsilon$ :

Given the above Lemmas and using the fact that for any two matrices $A,B$ we have $||AB||_\mathrm{op} \leq ||A||_\mathrm{op} ||B||_\mathrm{op}$ It is easy to see that for all terms $X$ except $B$, we have $||X||_\mathrm{op} \leq C\epsilon$ for some constant $C$. The only thing that we should calculate is the term $B$:
\begin{align}
    B = \frac{1}{k}\sum_{i=1}^k \Sigma (w^{(i)}_* - w_0)(w^{(i)}_* - w_0)^{\top} \Sigma = \Sigma \left(\frac{1}{k}\sum_{i=1}^k(w^{(i)}_* - w_0)(w^{(i)}_* - w_0)^{\top}\right) \Sigma.
\end{align}
By adding and subtracting terms, we have
\begin{align}
    B &= \Sigma \left(\frac{1}{k}\sum_{i=1}^k(w^{(i)}_* - w_0)(w^{(i)}_* - w_0)^{\top} -\Sigma^{-1} + \Sigma^{-1}\right) \Sigma
    \\
    &= \Sigma + \Sigma \left[\frac{1}{k}\sum_{i=1}^k(w^{(i)}_* - w_0)(w^{(i)}_* - w_0)^{\top} - \Sigma^{-1} \right] \Sigma
    \\
    &= \Sigma + \Sigma \underbrace{\left[\left(\frac{1}{k}\sum_{i=1}^k(w^{(i)}_* - w_0)(w^{(i)}_* - w_0)^{\top} - (H^{(i)})^{-1}\right) + \left(\frac{1}{k}\sum_{i=1}^k(H^{(i)})^{-1} -\Sigma^{-1}\right) \right]}_{\text{We need to show this term has } || \cdot ||_\mathrm{op} \lesssim \epsilon}  \Sigma
\end{align}
It's easy to see $||\frac{1}{k}\sum_{i=1}^k(H^{(i)})^{-1} -\Sigma^{-1}||_\mathrm{op} \lesssim \epsilon/d$ by triangle inequality, as each of $(H^{(i)})^{-1}$s are $\epsilon/d$ close to $\Sigma^{-1}$. Indeed, by using Lemma~\ref{spectrallemma} we can see that:
\begin{align}
    ||\frac{1}{k}\sum_{i=1}^k(H^{(i)})^{-1} -\Sigma^{-1}||_\mathrm{op} \leq 
    \frac{1}{k}\sum_{i=1}^k ||(H^{(i)})^{-1} -\Sigma^{-1}||_\mathrm{op} \leq \frac{2\epsilon}{d}
\end{align}
So it's enough to show that with high probability $||\frac{1}{k}\sum_{i=1}^k(w^{(i)}_* - w_0)(w^{(i)}_* - w_0)^{\top} - (H^{(i)})^{-1} ||_\mathrm{op} \lesssim \epsilon$.

Note that each $w_*^{(i)}$ is sampled from a distribution with mean $w_0$ and covariance matrix $(H^{(i)})^{-1}$. Based on Lemma~\ref{spectrallemma} we know all $(H^{(i)})^{-1}$s are close to $\Sigma^{-1}$, so we expect $\frac{1}{k}\sum_{i=1}^k(w^{(i)}_* - w_0)(w^{(i)}_* - w_0)^{\top}$ to be close to $\Sigma^{-1}$. If the covariance matrices were in fact equal, we could use the standard covariance concentration bounds (Lemmas~\ref{gausscov} and \ref{subgausscov}), but here, because they are not equal, we have to bound them some other way.
\qedhere
\end{continuedproof*}

Before proceeding with the proof, we state a Lemma that we need for the rest of the proof. The proof of this lemma can be found in \cite{vershynin2018high}.

\begin{auxproofbox}
\begin{lemma}\label{operator-covering}
    (Computing the operator norm on a net \cite{vershynin2018high}). Let $A$ be an $d \times d$ symmetric matrix and $\varepsilon \in [0,1)$. Then, for any $\varepsilon$-net $N_\epsilon$ of the sphere $S^{n-1}$, we have
\begin{align}
    \|A\|_\mathrm{op} \leq \frac{1}{1-2\varepsilon} \cdot \sup_{x \in N_\epsilon} |x^{\top}Ax|
\end{align}
\end{lemma}
\end{auxproofbox}

\begin{continuedproof}{Theorem~\ref{thm:main}}
To proceed with the proof, we use a covering argument. We consider $N_{1/4}$ a $1/4$-net of sphere $S^{n-1}$. By setting $A = \frac{1}{k}\sum_{i=1}^k(w^{(i)}_* - w_0)(w^{(i)}_* - w_0)^{\top} - (H^{(i)})^{-1}$ we have:
\begin{align}
    \|A\|_\mathrm{op} \leq \max_{x \in N_{1/4}} 2|x^{\top}Ax| , ~\text{s.t.}~
    A := \frac{1}{k}\sum_{i=1}^k(w^{(i)}_* - w_0)(w^{(i)}_* - w_0)^{\top} - (H^{(i)})^{-1}
\end{align}
For any $x \in N_{1/4}$ we have:
\begin{align}
    x^{\top}Ax &= x^{\top}\left[\frac{1}{k}\sum_{i=1}^k(w^{(i)}_* - w_0)(w^{(i)}_* - w_0)^{\top} - (H^{(i)})^{-1}\right]x 
    \\
    &= \frac{1}{k}\sum_{i=1}^k (x^{\top}(w^{(i)}_* - w_0))^2 - x^{\top}(H^{(i)})^{-1}x,
\end{align}
such that for each $i \in [k]$ we have:
\begin{align}
    \EX\left[(x^{\top}(w^{(i)}_* - w_0))^2 - x^{\top}(H^{(i)})^{-1}x\right] = x^{\top}\EX\left[(w^{(i)}_* - w_0)(w^{(i)}_* - w_0)\right]^{\top}x - x^{\top}(H^{(i)})^{-1}x = 0.
\end{align}
We can see that by Lemma~\ref{gaussian_is_sub} for each $i \in [k]$, we have $x^{\top}(w^{(i)}_* - w_0) \in SG_d(\|x\|_2^2\|(H^{(i)})^{-1}\|_\mathrm{op})$, and by Lemma~\ref{spectrallemma} and the fact that $\|x\|_2^2 \leq 1$ we can conclude $x^{\top}(w^{(i)}_* - w_0) \in SG(2/c)$ (e.g., $x^{\top}(w^{(i)}_* - w_0)$ is a sub-Gaussian vector with parameter $2/c$).

Now note that by Lemma~\ref{subgaussian-squared}, for every $i \in [k]$, $(x^{\top}(w^{(i)}_* - w_0))^2 - x^{\top}(H^{(i)})^{-1}x \in SE(128/c^2,4/c)$ and they are independent. By Lemma~\ref{subexp-compos} we have:
\begin{align}
    \frac{1}{k}\sum_{i=1}^k (x^{\top}(w^{(i)}_* - w_0))^2 - x^{\top}(H^{(i)})^{-1}x \in SE(128/c^2k,4/ck).
\end{align}
Note that by Theorem~\ref{covering-number}, we have $N_{1/4} \leq 9^d$. We can then use the standard sub-exponential concentration bound (Lemma~\ref{subexp-concentrate}) and union bound over all the $x \in N_{1/4}$:
\begin{align}
    &\mathbb{P}\left[||\frac{1}{k}\sum_{i=1}^k(w^{(i)}_* - w_0)(w^{(i)}_* - w_0)^{\top} - (H^{(i)})^{-1}||_\mathrm{op} \geq \epsilon\right] \leq
    \\
    &\mathbb{P}\left[\max_{x \in N_{1/4}} \mid x^{\top}\left(\frac{1}{k}\sum_{i=1}^k(w^{(i)}_* - w_0)(w^{(i)}_* - w_0)^{\top} - (H^{(i)})^{-1}\right)x \mid \geq \epsilon/2\right] \leq
    \\
    & 9^d \mathbb{P}\left[SE(128/c^2k,4/ck) \geq \epsilon/2 \right] \leq
    \\
    & \exp\left(-\frac{1}{2} \min\left\{\frac{c^2 k \epsilon^2}{512}, \frac{c k \epsilon}{8}\right\} + 3d \right) \leq
    \\
    & \exp\left(-\frac {c^2 k \epsilon^2}{1024} + 3d \right),
\end{align}
where the first inequality is by using the net argument, the second inequality is by union bounding and the fact that random variables are sub-exponential, the third inequality is by probability bound of sub-exponential random variables, and the last inequality is by simple algebraic manipulations. We want this event  (which we call $\mathcal{E}_4$) to happen with probability at most $\delta$. Hence, we have:
\begin{align}
    \exp\left(-\frac {c^2 k \epsilon^2}{1024} + 3d \right) \leq \delta \Leftrightarrow 
    k \geq \frac{1024(\log(1/\delta) + 3d)}{c^2 \epsilon^2}.
\end{align}
Hence, ignoring absolute constants, it is enough to have a large number of batches $k$, i.e.,
\begin{align}
    k \gtrsim \frac{d + \log(1/\delta)}{\epsilon^2}.
\end{align}
For all of our results to fail, at least one of the events 
$\mathcal{E}_1$, $\mathcal{E}_2$, $\mathcal{E}_3$, or $\mathcal{E}_4$ must fail. 
Recall that the validity of $\mathcal{E}_2$ and $\mathcal{E}_3$ depends on the success 
of $\mathcal{E}_1$, while the validity of $\mathcal{E}_4$ depends on the joint success 
of $\mathcal{E}_1$, $\mathcal{E}_2$, and $\mathcal{E}_3$. For any event $\mathcal{E}_i$, 
we denote its complement by $\bar{\mathcal{E}}_i$. Thus, because we bound a finite number of matrices to have operator norm $\leq C\epsilon$ for some constant $C$, for some absolute constant $C'$ we obtain:
\begin{align}
    \mathbb{P}\left[ \|S_g(w) - \Sigma \|_\mathrm{op} > C' \epsilon \right] &\leq
    \mathbb{P}\left[\bar{\mathcal{E}_1} \vee  \bar{\mathcal{E}_2} \vee  \bar{\mathcal{E}_3} \vee  \bar{\mathcal{E}_4}  \right] 
    \\
    &\leq \mathbb{P}[\bar{\mathcal{E}_1}] + \mathbb{P}[\mathcal{E}_1] \mathbb{P}[\bar{\mathcal{E}_2} \mid \mathcal{E}_1] + \mathbb{P}[\mathcal{E}_1]\mathbb{P}[\bar{\mathcal{E}_3} \mid \mathcal{E}_1] + \mathbb{P}[\mathcal{E}_1 , \mathcal{E}_2 , \mathcal{E}_3] \mathbb{P}[\bar{\mathcal{E}_4} \mid \mathcal{E}_1 , \mathcal{E}_2 , \mathcal{E}_3]
    \\
    & \leq \mathbb{P}[\bar{\mathcal{E}_1}] + \mathbb{P}[\bar{\mathcal{E}_2} \mid \mathcal{E}_1] + \mathbb{P}[\bar{\mathcal{E}_3} \mid \mathcal{E}_1] + \mathbb{P}[\bar{\mathcal{E}_4} \mid \mathcal{E}_1 , \mathcal{E}_2 , \mathcal{E}_3] 
    \\
    & \leq 4 \delta
\end{align}
\end{continuedproof}
\subsection{Proof of Lemma~\ref{lemma:smallbatch}}
\begin{proof}
We compute $\mathrm{Cov}(\nabla L^{(j)}(w))$ for the $j$'th batch. As the distribution for all batches is equal, for simplicity of the notation, we remove the batch index $j$ in this proof. We define $H = \frac{1}{n} X^\top X = \frac{1}{n} \sum_{i=1}^n x_ix_i^{\top}$. First note that 
\begin{align}
    \nabla L(w) &= \frac{1}{n} X^\top (Xw - y) = \frac{1}{n} X^\top (Xw - Xw_0 - \varepsilon) 
    \\
    &= \frac{1}{n} X^\top X w- \frac{1}{n} X^\top X w_0 - \frac{1}{n} X^\top \varepsilon =  H(w - w_0) - \frac{1}{n}\sum_{i=1}^n \varepsilon_ix_i
\end{align}
For simplicity we call $\frac{1}{n}\sum_{i=1}^n \varepsilon_ix_i = \zeta$. Hence we have 
\begin{align}
    \EX[\nabla L(w)] = \EX[H] (w-w_0) - \sum_{i=1}^n \EX[\varepsilon_i]\EX[x_i] = \Sigma (w-w_0)
\end{align}
where the second expectation is zero because $\varepsilon_i$ and $x_i$ are independent. Then we have:
\begin{align}
    \mathrm{Cov}(\nabla L(w)) =& \EX[(\nabla L(w) - \EX[\nabla L(w)])(\nabla L(w) - \EX[\nabla L(w)])^{\top}]
    \\
    =& \EX[((H - \Sigma)(w-w_0) -  \zeta) ((H - \Sigma)(w-w_0) -  \zeta)^{\top}]
\end{align}
which is equal to
\begin{align}
        \mathrm{Cov}(\nabla L(w)) =& \EX[\zeta\zeta^{\top} - ((H - \Sigma)(w-w_0))\zeta^{\top} - \zeta((H - \Sigma)(w-w_0))^{\top} 
        \\
        +& (H - \Sigma)(w-w_0)(w-w_0)^{\top}(H-\Sigma)]
\end{align}
There are four terms, and we derive them individually.

1- $\EX[\zeta\zeta^{\top}]$:
\begin{align}
    \EX[\zeta\zeta^{\top}] &= \frac{1}{n^2}\sum_{i=1}^n\sum_{j=1}^n\EX [\varepsilon_i\varepsilon_jx_ix_j^{\top}] 
    \\
    &= \frac{1}{n^2} \sum_{i \neq j} \EX \varepsilon_i \EX \varepsilon_j \EX x_i \EX x_j^{\top} + \frac{1}{n^2} \sum_{i=1}^n \EX[\varepsilon_i\varepsilon_i] \EX[x_ix_i^{\top}]
    \\
    &= 0 + \frac{1}{n^2}\sum_{i=1}^n \sigma^2 \EX[x_ix_i^{\top}] = \frac{\sigma^2}{n}\Sigma.
\end{align}
2- $\EX[((H - \Sigma)(w-w_0))\zeta^{\top}]$:
\begin{align}
    \EX[((H - \Sigma)(w-w_0))\zeta^{\top}] =& \frac{1}{n}\sum_{i=1}^n \EX[H(w-w_0)\varepsilon_ix_i^{\top}] 
    \\
    -& \frac{1}{n}\sum_{i=1}^n \EX[\Sigma(w-w_0)\varepsilon_ix_i^{\top}]
    \\
    =& \frac{1}{n}\sum_{i=1}^n \EX_X[\EX_\varepsilon[H(w-w_0)\varepsilon_ix_i^{\top} | X]] 
    \\
    -& \frac{1}{n}\sum_{i=1}^n \EX_X[\EX_\varepsilon[\Sigma(w-w_0)\varepsilon_ix_i^{\top} | X]] = 0.
\end{align}
Both expectation terms are zero because the inner expectations are zero.

3- $\EX[\zeta((H - \Sigma)(w-w_0))^{\top}]$: Similarly to case 2, we can write
\begin{align}
    \EX[\zeta((H - \Sigma)(w-w_0))^{\top}] =& \frac{1}{n}\sum_{i=1}^n \EX[\varepsilon_ix_i(w-w_0)^{\top}H] 
    \\
    -& \frac{1}{n}\sum_{i=1}^n \EX[\varepsilon_ix_i(w-w_0)^{\top}\Sigma]
    \\
    =& \frac{1}{n}\sum_{i=1}^n \EX_X[\EX_\varepsilon[\varepsilon_ix_i(w-w_0)^{\top}H | X]] 
    \\
    -& \frac{1}{n}\sum_{i=1}^n \EX_X[\EX_\varepsilon[\varepsilon_ix_i(w-w_0)^{\top}\Sigma | X]] = 0.
\end{align}
4- $\EX[(H - \Sigma)(w-w_0)(w-w_0)^{\top}(H-\Sigma)]$: for simplicity we call $u = w-w_0$. We use $u$ instead of $w-w_0$ for clear notation. We have:
\begin{align}
    \EX[(H - \Sigma)uu^{\top}(H-\Sigma)] =& \EX[\frac{1}{n}(\sum_{i=1}^n x_ix_i^{\top} - \Sigma)uu^{\top}(\sum_{i=1}^n x_ix_i^{\top} - \Sigma)\frac{1}{n}]
    \\
    =& \frac{1}{n^2} \sum_{i \neq j} (\EX(x_ix_i^{\top}) - \Sigma)uu^{\top}(\EX(x_jx_j^{\top}) - \Sigma)
    \\
    +& \frac{1}{n^2} \sum_{i=1}^n \EX[(x_ix_i^{\top} - \Sigma)uu^{\top}(x_ix_i^{\top} - \Sigma)]
    \\
    =& \frac{1}{n}\EX[(x_1x_1^{\top} - \Sigma)uu^{\top}(x_1x_1^{\top} - \Sigma)].
\end{align}
We drop the subscript $1$ from now on in this proof and will use a subscript for the index. We denote the $j$'th index of $x$ with $x_{[j]} \in \mathbb{R}$ in the rest of the proof. We continue to derive $\EX[(xx^{\top} - \Sigma)uu^{\top}(xx^{\top} - \Sigma)].$
\begin{align}
    \EX[(xx^{\top} - \Sigma)uu^{\top}(xx^{\top} - \Sigma)] =& \Sigma u u^{\top}\Sigma -\EX[xx^{\top}]uu^{\top}\Sigma 
- \Sigma uu^{\top}\EX[xx^{\top}] + \EX[xx^{\top}uu^{\top}xx^{\top}]
    \\
    =& \EX[xx^{\top}uu^{\top}xx^{\top}] - \Sigma uu^{\top} \Sigma.
\end{align}

Note that because we assumed $\Sigma \preceq I$ we have:
\begin{align}
    \|\Sigma uu^{\top} \Sigma\|_\mathrm{op} \leq \|\Sigma\|_\mathrm{op}^2 \|u\|_2^2 = \|\Sigma\|_\mathrm{op}^2 \|w - w_0\|_2^2 \leq \|w - w_0\|_2^2.
\end{align}

Now we proceed to bound the operator norm of  $\EX[xx^{\top}uu^{\top}xx^{\top}] = \EX[(x^{\top}u)^2xx^{\top}]$. Note that this matrix is symmetric and positive semi-definite. Now observe that for any $v \in \mathbb{R}^d$ such that $\|v\|_2 = 1$ we have:
\begin{align}
    v^\top \EX[(x^{\top}u)^2xx^{\top}] v =
    \EX[(x^{\top}u)^2 v^\top xx^{\top}v] = \|u\|_2^2 \mathbb{E}[(x^{\top}u/\|u\|_2)^2 (x^{\top}v)^2] \leq \|u\|_2^2 \sqrt{\mathbb{E}[(x^{\top}u/\|u\|_2)^4]} \sqrt{\mathbb{E}[(x^{\top}v)^4]},
\end{align}
where the inequality is by Cauchy-Schwarz. Now note that because $x \in \mathrm{SG}_d(1)$ by our assumption, both $x^{\top}u/\|u\|_2$ and $x^{\top}v$ are $\mathrm{SG}(1)$ by definition. Hence by using the bound on $m$-th moment of sub-Gaussian random variables with $m = 4$ (\cite{vershynin2018high}), we have:
\begin{align}
    \|u\|_2^2\sqrt{\mathbb{E}[(x^{\top}u/\|u\|_2)^4]} \sqrt{\mathbb{E}[(x^{\top}v)^4]} \leq C'\|u\|_2^2  = C'\|w-w_0\|_2^2 
\end{align}
for some absolute constant $C'$ and by using the definition of $u = w - w_0$. Hence, by the triangle inequality, we have:
\begin{align}
\|\EX[xx^{\top}uu^{\top}xx^{\top}] - \Sigma uu^{\top} \Sigma\|_\mathrm{op} \leq \|\EX[xx^{\top}uu^{\top}xx^{\top}]\|_\mathrm{op} + \|\Sigma uu^{\top} \Sigma\|_\mathrm{op} \leq (C'+1) \|w-w_0\|_2^2 = C\|w-w_0\|_2^2,
\end{align}
for $C = C' + 1$.
Now by setting $\mathrm{noise} = \EX[xx^{\top}uu^{\top}xx^{\top}] - \Sigma uu^{\top} \Sigma$ we have:
\begin{align}
\mathrm{Cov}(\nabla L^{(j)}(w)) = \frac{\sigma^2}{n} \Sigma + \frac{\mathrm{noise}}{n},
\end{align}
such that 
\begin{align}
    \|\mathrm{noise}\|_\mathrm{op} \leq C \|w - w_0\|_2^2,
\end{align}
for some absolute constant $C$.
\end{proof}
\subsection{Proof of Lemma~\ref{thm:expected}}
\begin{proof}
    The first part of the proof is identical to the first part of the proof of Lemma~\ref{lemma:smallbatch}, in which we have:
    \begin{align}
        \mathrm{Cov}(\nabla L^{(j)}(w)) = \frac{\sigma^{2}}{n}\,\Sigma 
\;+\; \frac{1}{n}\,\bigl[\EX[xx^{\top}(w-w_0)(w-w_0)^{\top}xx^{\top}] - \Sigma (w-w_0)(w-w_0)^{\top} \Sigma]
    \end{align}
    The only difference is that in this case $x \sim \mathcal{N}(0,\Sigma) $, hence we can derive $\EX[xx^{\top}(w-w_0)(w-w_0)^{\top}xx^{\top}] = \EX[(x^{\top}(w-w_0))^2xx^{\top}]$ explicitly. For simplicity, as before, we define $u := w-w_0$.

    We call the matrix $\EX[(x^{\top}u)^2xx^{\top}] := A$. Now we proceed to compute the value of $A_{[i,j]}$ for $i,j \in [d].$
\begin{align}
    A_{[i,j]}= \EX[(\sum_{k=1}^d x_{[k]}u_{[k]})^2x_{[i]}x_{[j]}] = \sum_{k=1}^d \sum_{p=1}^d u_{[k]} u_{[p]}\EX[x_{[k]}x_{[p]}x_{[i]}x_{[j]}].
\end{align}

To derive the fourth moment, we use Isserlis's theorem~(Theorem~\ref{Isserlis}). Note that $x_{[k]}x_{[p]}x_{[i]}x_{[j]}$ are jointly Gaussian, even if some of the indices get repeated.
\begin{align}
    \EX[x_{[k]}x_{[p]}x_{[i]}x_{[j]}] &= \EX[x_{[k]} x_{[p]}]\EX[x_{[i]}x_{[j]}] + \EX[x_{[k]}x_{[i]}]\EX[x_{[p]}x_{[j]}] + \EX[x_{[k]}x_{[j]}]\EX[x_{[p]}x_{[i]}]
    \\
    &= \Sigma_{[k,p]}\Sigma_{[i,j]} + \Sigma_{[k,i]}\Sigma_{[p,j]}+ \Sigma_{[k,j]}\Sigma_{[p,i]}.
\end{align}
Hence, we have:
\begin{align}
    A_{[i,j]} &= \sum_{k=1}^d \sum_{p=1}^d u_{[k]} u_{[p]} (\Sigma_{[k,p]}\Sigma_{[i,j]} + \Sigma_{[k,i]}\Sigma_{[p,j]}+ \Sigma_{[k,j]}\Sigma_{[p,i]})
    \\
    &=  \sum_{k=1}^d \sum_{p=1}^d u_{[k]} u_{[p]} \Sigma_{[k,p]}\Sigma_{[i,j]} + 
    \sum_{k=1}^d \sum_{p=1}^d u_{[k]} u_{[p]} \Sigma_{[k,i]}\Sigma_{[p,j]} + 
    \sum_{k=1}^d \sum_{p=1}^d u_{[k]} u_{[p]} \Sigma_{[k,j]}\Sigma_{[p,i]}
    \\
    &= (u^{\top}\Sigma u) \Sigma_{[i,j]} + 2(\sum_{k=1}^d \Sigma_{[i,k]} u_{[k]})(\sum_{k=1}^d \Sigma_{[j,k]} u_{[k]}).
\end{align}
where in the last line we used the symmetric property of $\Sigma$
Now note that $(\sum_{k=1}^d \Sigma_{[i,k]} u_{[k]})(\sum_{k=1}^d \Sigma_{[j,k]} u_{[k]}) = (\Sigma uu^{\top} \Sigma)_{[i,j]}$ and $(u^{\top}\Sigma u) \Sigma_{[i,j]} = ((u^{\top}\Sigma u) \Sigma)_{[i,j]}$.

By combining all parts and the fact that $u = w - w_0$ we have:
\begin{align}
    \mathrm{Cov}(\nabla L^{(j)}(w)) = \frac{\sigma^{2}}{n}\,\Sigma 
\;+\; \frac{1}{n}\,\bigl[
     \Sigma\,(w - w_0)(w - w_0)^{\mathsf T}\Sigma 
     \;+\;
     \bigl((w - w_0)^{\mathsf T}\Sigma (w - w_0)\bigr)\Sigma
\bigr].
\end{align}
\end{proof}
\subsection{Proof of Corollary~\ref{cor:necessary}}
\begin{proof}
    For any $w \in \mathbb{R}^d$ we have:
    \begin{align}
         \|\mathrm{Cov}(\nabla L^{(j)}(w)) - \Sigma\|_\mathrm{op}
         &=
         \|\frac{\sigma^{2}}{n}\,\Sigma 
\;+\; \frac{1}{n}\,\bigl[
     \Sigma\,(w - w_0)(w - w_0)^{\mathsf T}\Sigma 
     \;+\;
     \bigl((w - w_0)^{\mathsf T}\Sigma (w - w_0)\bigr)\Sigma\bigr] - \Sigma\|_\mathrm{op}
     \\
     &= \|(\frac{\sigma^2}{n} - 1 + \frac{1}{n}(w - w_0)^{\top}\Sigma (w - w_0))\Sigma + \frac{1}{n}\,
     \Sigma\,(w - w_0)(w - w_0)^{\mathsf T}\Sigma 
      \|_\mathrm{op}.
    \end{align}
Now note that as $\Sigma$ is positive definite, for any $C >0$ we can find at least one $w$ (we call it $w_C$) far from $w_0$ such that it ensures
\begin{align}
    \frac{\sigma^2}{n} - 1 + \frac{1}{n}(w_C - w_0)^{\top}\Sigma (w_C - w_0) > C.
\end{align}
Now note that we can see that for any $w \in \mathbb{R}^d, v \in \mathbb{R}^d: v \neq 0$ we have
\begin{align}
    v^\top\Sigma\,(w - w_0)(w - w_0)^{\mathsf T}\Sigma v = ( v^\top\Sigma\,(w - w_0))^2 \geq 0,
\end{align}
meaning $\Sigma\,(w - w_0)(w - w_0)^{\mathsf T}\Sigma$ is positive semidefinite. Hence for any $C > 0$ we have:
\begin{align}
    \|\mathrm{Cov}(\nabla L^{(j)}(w_C)) - \Sigma\|_\mathrm{op}
      &=
      \|(\frac{\sigma^2}{n} - 1 + \frac{1}{n}(w_C - w_0)^{\top}\Sigma (w_C - w_0))\Sigma + \frac{1}{n}\,
     \Sigma\,(w_C - w_0)(w_C - w_0)^{\mathsf T}\Sigma 
      \|_\mathrm{op}
      \\
      &\geq
      \|(\frac{\sigma^2}{n} - 1 + \frac{1}{n}(w_C - w_0)^{\top}\Sigma (w_C - w_0))\Sigma\|_\mathrm{op} 
      \\
      &\geq
      C \|\Sigma\|_\mathrm{op}.
\end{align}
Letting \( C \to \infty \) completes the proof.
\end{proof}
\subsection{Proof of Lemma~\ref{lem:ctrexample}}
\begin{proof}
    Suppose $w_0 = 1, n =1 , \sigma^2 = 1$ and suppose $\mathcal{D}_x$ is the Rademacher random variable, i.e., $\mathbb{P}(x = 1) = \mathbb{P}(x = -1) = 1/2$. Hence based on the proof of Lemma~\ref{lemma:smallbatch} we have:
    \begin{align}
        \mathrm{Cov}(\nabla L^{(j)}(w)) =  \Sigma + (w - 1)^2\mathbb{E}[x^4] - (w - 1)^2
        \Sigma^2.
    \end{align}
    Now note that the variance of the Rademacher random variable is $1$, so we have:
    \begin{align}
        \mathrm{Cov}(\nabla L^{(j)}(w)) =  \Sigma + (w - 1)^2[\mathbb{E}[x^4] - 1] = \Sigma + (w - 1)^2[1 - 1] = \Sigma,
    \end{align}
    where in the last step we used the simple fact that $\mathbb{E}[x^4] = \mathbb{E}[1] = 1$.
\end{proof}
\subsection{Proof of Theorem~\ref{theorem:precond}}
\begin{proof} Before proving the result note that using a similar argument as Lemma~\ref{weyl}, we can infer closeness of $(S_g(w))^{-1}$ and $\Sigma^{-1}$ by closeness of $S_g(w)$ and $\Sigma$, given sufficient regularity conditions ($\lambda_d(\Sigma)$ being a constant and $\epsilon < \lambda_d(\Sigma)/2$). Hence,  $(S_g(w))^{-1}$ and $\Sigma^{-1}$ being close is a natural assumption given the results of Section~\ref{sec:non-asymp}.

Suppose $w_*$ is the empirical optimal parameter. For simplicity, we use gradient descent with a step size of 1. Note that:
\begin{align}
        w_{t+1} - w_* &= w_t - (S_g(w_t))^{-1}\Sigma(w_t - w_*) - w_*
        \\
        &= w_t - (\Sigma^{-1} + (S_g(w_t))^{-1} - \Sigma^{-1})\Sigma(w_t - w_*) - w_*
        \\
        &=  ((S_g(w_t))^{-1} - \Sigma^{-1})\Sigma(w_t - w_*).
    \end{align}
    Hence, we have
    \begin{align}
        \|w_{t+1} - w_*\|_2 = \|((S_g(w_t))^{-1} - \Sigma^{-1})\Sigma(w_t - w_*)\|_2 \leq \epsilon \|w_t - w_*\|_2.
    \end{align}
    Where the inequality is by simply using $\|AB\|_\mathrm{op} \leq \|A\|_\mathrm{op} \|B\|_\mathrm{op}$ and the fact that $\Sigma \preceq I$. Then, by induction, we have
    \begin{align}
        \|w_{t+1} - w_*\|_2 \leq \epsilon^t \|w_1 - w_*\|_2,
    \end{align}
    and the convergence rate will follow from this.
\end{proof}
\subsection{Proof of Lemma~\ref{lemma:advrisk}}
\begin{proof}
    The proof relies on the fact that
    \begin{align}
        \|w - w_0\|^2_{\Sigma} &= (w-w_0)^\top \Sigma \Sigma^{-1}\Sigma (w-w_0) \\
        &= (\Sigma(w-w_0))^\top \Sigma^{-1}(\Sigma(w-w_0)) 
    \end{align}
    The rest of the proof follows using the approximations.
\end{proof}

\subsection{Sub-Exponential Property of Batch Gradients}\label{theory:subexp}
Note that, as before, for any batch $j \in [k]$ we have:
\begin{align}
    \nabla L^{(j)}(w) &= \frac{1}{n}\sum_{i=1}^n x_i x_i^\top(w - w_0) - \frac{1}{n}\sum_{i=1}^n \varepsilon_ix_i.
    \\
    &= \frac{1}{\sqrt{n}}\sum_{i=1}^n x_i x_i^\top(w - w_0)/
    \sqrt{n}- \frac{1}{\sqrt{n}}\sum_{i=1}^n \varepsilon_ix_i / \sqrt{n}.
\end{align}
where $x_i$ is the $i$th data point of batch $j$ and $\varepsilon_i \sim \mathcal{N}(0,n)$.
Now for any $v \in \mathbb{R}^d$ such that $\|v\|_2 = 1$ we have
\begin{align}
    v^\top \nabla L^{(j)}(w) = \frac{1}{\sqrt{n}}\sum_{i=1}^n (v^\top x_i) (x_i^\top(w - w_0)/\sqrt{n}) - \frac{1}{\sqrt{n}}\sum_{i=1}^n  (\varepsilon_i/\sqrt{n}) (v^\top x_i).
\end{align}
Now note that in both large batch size and small batch size regimes, we have $n = \Omega(\|w-w_0\|_2^2)$. Hence we have that $\sqrt{n} = \Omega(\|w-w_0\|_2)$ and by the fact that $x_i \sim \mathrm{SG}_d(1)$, each of the $(\cdot)$ terms are $\mathrm{SG(1)}$, hence by usual properties of sub-Gaussian and sub-exponential random variables \cite{vershynin2018high, wainwright2019high} (e.g., product of two dependent sub-Gaussian is sub-exponential, sum of independent sub-exponentials is sub-exponential, and sum of dependent sub-exponentials is sub-exponential) we can see that the whole expression will be $\mathrm{SE(\mathcal{O}(1), \mathcal{O}(1))}$.

\section{PRACTICAL CONSIDERATIONS}
\label{sec:practical}

\subsection{Gradient Covariance Inversion}\label{sec:inversion}
As discussed in the paper, many applications, such as preconditioning, require 
computing the action of the inverse empirical gradient covariance on a vector, 
$(S_g(w))^{-1} v$. This can be done efficiently by solving the linear system
\begin{align}
S_g(w) u = v,
\end{align}
via a least-squares method, rather than explicitly forming the inverse matrix.

\subsection{Scaling}
\label{sec:scaling}
While we assumed $x \in \mathrm{SG}_d(1)$, which implies $\Sigma \preceq I$ by standard 
properties of sub-Gaussian vectors, all arguments extend to the more general case 
$x \in \mathrm{SG}_d(\alpha)$, yielding $\Sigma \preceq \alpha I$ for some $\alpha > 1$. 
For instance, in Lemma~\ref{lemma:smallbatch} the noise term satisfies  
\begin{align}
\|\mathrm{noise}/n\|_\mathrm{op} \leq C \alpha^2 \|w-w_0\|_2^2 / n,
\end{align}
so achieving 
\begin{align}
\|\mathrm{noise}/n\|_\mathrm{op} \leq \epsilon \|\Sigma\|_\mathrm{op}
\end{align}
requires  
\begin{align}
n = \Omega\!\left(\frac{\alpha \|w-w_0\|_2^2}{\epsilon}\right).
\end{align}

The lesson here is that the error term is not scale-independent: its magnitude depends 
on the sub-Gaussian parameter $\alpha$, which in turn affects the required batch size.

\subsection{Another Estimator to Consider}
Another possible estimator is obtained by adding noise $\mathcal{N}(0,\alpha)$ to the 
targets and using $nS_g(w)/\alpha$ as the estimator. If the targets already contain 
inherent noise $\mathcal{N}(0,\widetilde{\sigma}^2)$, then by 
Lemma~\ref{lemma:smallbatch} one can see that letting $\alpha \to \infty$ makes the 
estimator converge to $\Sigma$. However, this comes at a cost: as $\alpha \to \infty$, the information about the mean gradient is effectively destroyed. In particular, each batch gradient becomes
\begin{align}
    \nabla L^{(j)}(w) &= \frac{1}{n}\sum_{i=1}^n x_i x_i^\top (w - w_0) 
    - \frac{1}{n}\sum_{i=1}^n \mathcal{N}(0,\widetilde{\sigma}^2+\alpha)_i\, x_i,
\end{align}
where the fluctuations in the second term are unrelated to the true gradient 
$\Sigma(w-w_0)$, dominate the first term. This raises two points:
\begin{enumerate}
    \item If we have access to two sets of gradients for each batch, one with clean targets and the other with noisy targets, we can use the former to approximate the gradient and the latter to approximate the Hessian.
    \item If we only have access to a single gradient per batch, is there a way to remove the effect of the added noise when estimating the mean gradient? In other words, can we perform post hoc variance reduction to largely undo the additional variance?
\end{enumerate}

\section{ADDITIONAL EXPERIMENTS}

\subsection{Data generation from Gaussian distribution}
\label{sec:data_gen}
In the experiments with data generated from a Gaussian distribution, we first sample a matrix of features from a multivariate Gaussian distribution and then pass the obtained features through a linear regression model to calculate the target values $y$. We add a Gaussian noise with a standard deviation of 0.1 to the target values to mimic the inherent measurement noise in real-world datasets. In the experiments, we refer to these targets as `clean,' while the `noisy' targets have an additional noise with variance equal to the batch size. In Section~\ref{sec:exp_gen_data} and Appendix~\ref{sec:exp_gen_data_diag}, the number of generated features is set to 6.

\subsection{Covariance estimate with a diagonal matrix}
\label{sec:exp_gen_data_diag}

\begin{figure}[!t]
    \centering
    \includegraphics[width=0.8\linewidth]{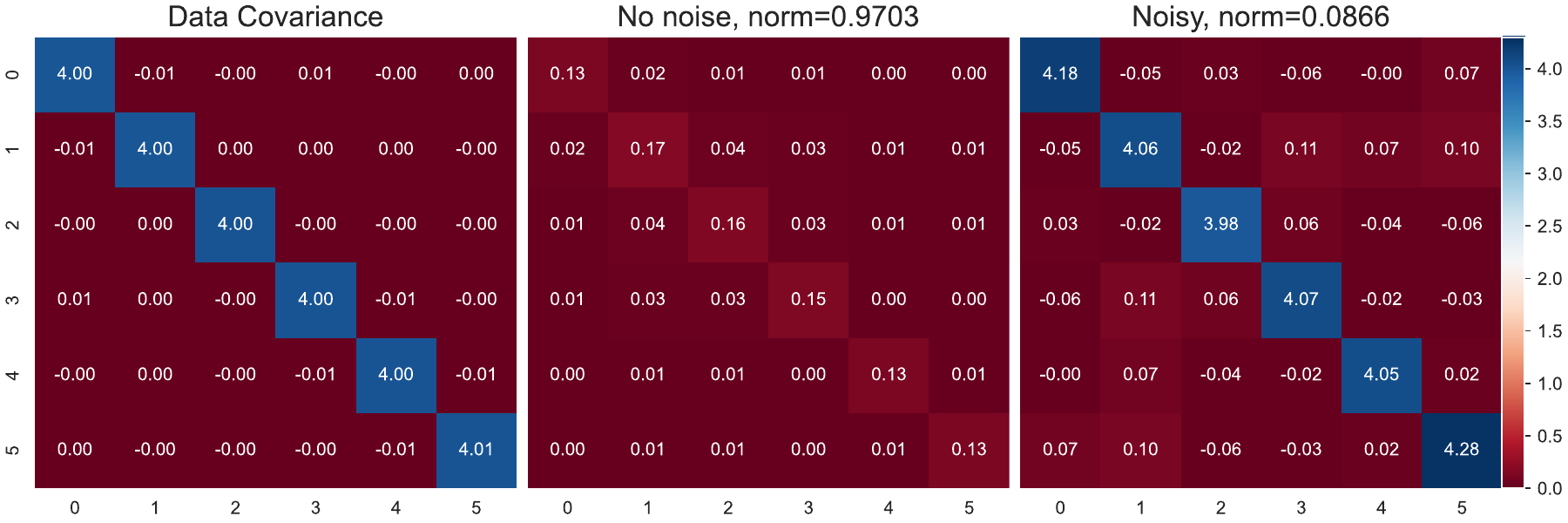}

    \caption{\textbf{Comparison of true data covariance and its estimates for the data generated from a Gaussian distribution with a diagonal covariance matrix.} Data covariance (left), its estimates by gradient covariance matrices obtained with no noise in the targets (middle), and with target noise of variance equal to batch size, $n=256$ (right). We compare the results using an operator norm in Eq.~\ref{eq:norm_r}. }
    \label{fig:illustration_matrices_diag}
\end{figure}

Here, we compare the covariance matrix estimates in the case of data generated from a Gaussian distribution with a diagonal covariance matrix, where the variance is 4 in both the clean target and noisy target cases (see Fig.~\ref{fig:illustration_matrices_diag}). Similar to Section~\ref{sec:exp_gen_data}, we used batch size (and the noise variance) $n=256$ with $k=3125$.

\subsection{Selection of batch size}

\label{sec:batch_size_select}

\begin{figure}[!t]
    \centering
    \includegraphics[width=
    \linewidth]{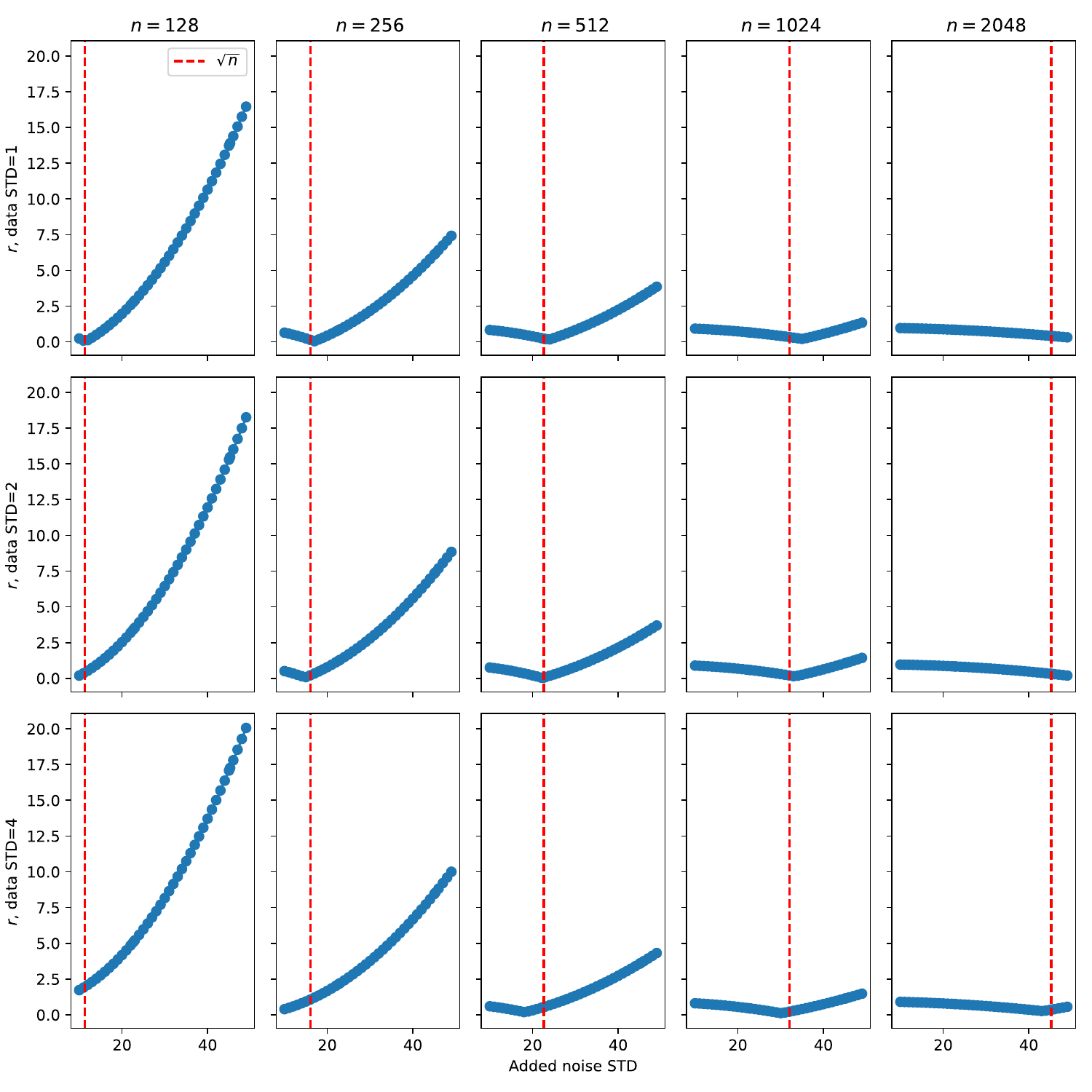}
    \caption{\textbf{Effects of the selected batch size and added noise on the estimate of the data covariance for the datasets coming from a Gaussian distribution with a diagonal covariance matrix with various standard deviations.} Rows: the results for the datasets with standard deviation equal to 1 (top), 2 (middle) and 4 (bottom). Columns:  the results for batch sizes $n$ equal to 128, 256, 512, 1024, and 2048 (from left to right). For all plots, the estimates of the covariance matrix are calculated with the standard deviation of noise added to the targets that varies from 10 to 50. The target noise standard deviation that is equal to $\sqrt{n}$ is marked with a vertical red dotted line. For all three datasets (rows), the sufficiently large batch size and the added target noise of $\sqrt{n}$ provide good estimates of the true data covariance.}
    \label{fig:sensitivity_batch_size_std}
\end{figure}

The proposed method assumes the selection of a batch size as a hyperparameter. 
In this evaluation, we demonstrate the effects of batch size values and the corresponding added target noise on three generated datasets.
We generated three datasets of 100K samples from a Gaussian distribution with a mean of zero and a diagonal covariance matrix with standard deviations of 1, 2, and 4. In this experiment, the generated data has two features. 
For each dataset, we calculated the relative operator norm $r$ for the data covariance estimates with the gradient covariance matrix testing batch sizes $n$ equal to 128, 256, 512, 1024, and 2048. For each batch size, the standard deviation of noise added to the targets was changed from 10 to 50. The results of the runs are summarised in Fig.~\ref{fig:sensitivity_batch_size_std}.
We notice that the datasets with a larger standard deviation benefit more from larger batch sizes $n$. See Appendix~\ref{sec:scaling} for the discussion. We also notice that the results with larger batch sizes are substantially more robust to changes in the added noise level in the targets. The noise variance, equal to the batch size, provides a reasonable estimate; however, for larger batch sizes, other values also provide good results.

\subsection{Data preprocessing and training details for public datasets}
\label{sec:data_preproc}

In this section, we provide experimental details for the runs on four public datasets. 

The wave energy converters dataset stores the values for the whole region in Watts as units. We divided the target variable by $10^6$ to work with megawatts instead of watts. Additionally, we rescaled the position columns by division by 100 and the power columns by $10^4$. For the California housing dataset, the `Population' feature column was divided by 1000. For the bike sharing dataset, the target variable was transformed to a logarithmic scale.
Both wine quality and bike sharing datasets exhibit correlated and redundant input features. To improve the model stability, we removed the `density' feature from the wine quality dataset and features `dteday', `atemp', `windspeed', and `workingday' from the bike sharing dataset in the pre-processing step. For the wine dataset we divided `total\_sulfur\_dioxide' and `free\_sulfur\_dioxide' by 100. We represented categorical features with one-hot encodings. For each dataset, 10\% of the data was reserved as a test set, the training set is used for training, and a validation set is provided for hyperparameter selection. We repeated the experiments with 10 random seeds, namely [0, 10, 20, 30, 40, 50, 60, 70, 80, 90].

\textbf{Linear regression} We trained all models with the Adam optimiser \citep{kingma2014adam}. We used the following number of epochs and learning rates: for the bike sharing data, 50 epochs with a learning rate of 0.0025, for the California housing data, 300 epochs with a learning rate of 0.0001, and for the wine quality data, 100 epochs with a learning rate of 0.001. For wave energy converters, we trained the model for 25 epochs with a learning rate of $10^{-5}$ and then for another 25 epochs with a learning rate of $10^{-6}$ and then $10^{-7}$. The batch size of 64 was used for all datasets considered in the experiment.

\textbf{Multi-layer perceptron} We trained all models with the Adam optimiser \citep{kingma2014adam}, similar as in the case of linear models. We used the following number of epochs and learning rates: for the bike sharing data, 50 epochs with a learning rate of 0.0005, for the California housing data, 100 epochs with a learning rate of 0.0001, and for the wine quality data, 300 epochs with a learning rate of 0.001. For wave energy converters, we trained the model for 25 epochs with a learning rate of $10^{-5}$ and then for another 25 epochs with a learning rate of $10^{-6}$ and then $10^{-7}$. The batch size of 64 was used for all datasets.

\subsection{Hessians of MLPs with ReLU}
\label{sec:hessian_relu_mlp}

\subsubsection{Structure of the Hessian matrix}
\begin{figure}[!t]
    \centering
    \includegraphics[width=1.0\linewidth, trim={0 0 0 0}, clip]{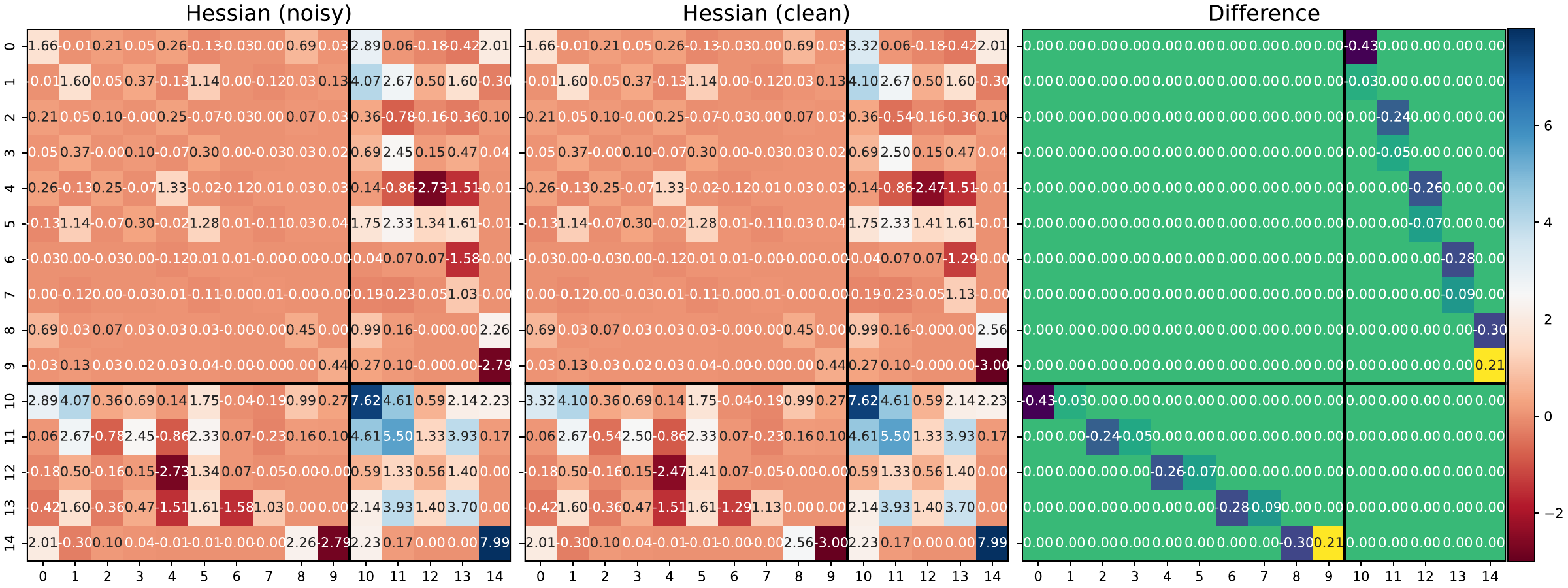}
    \includegraphics[width=1.0\linewidth, trim={0 0 0 0}, clip]{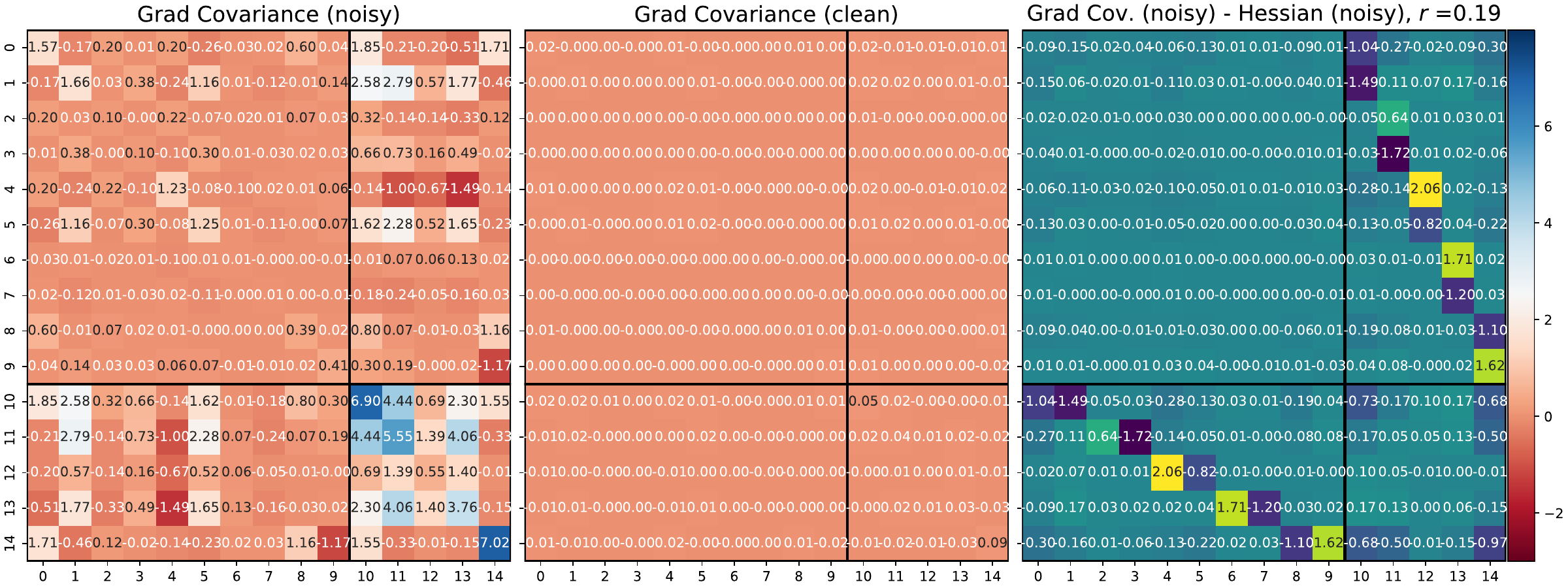}
    \includegraphics[width=1.0\linewidth, trim={0 0 0 0}, clip]{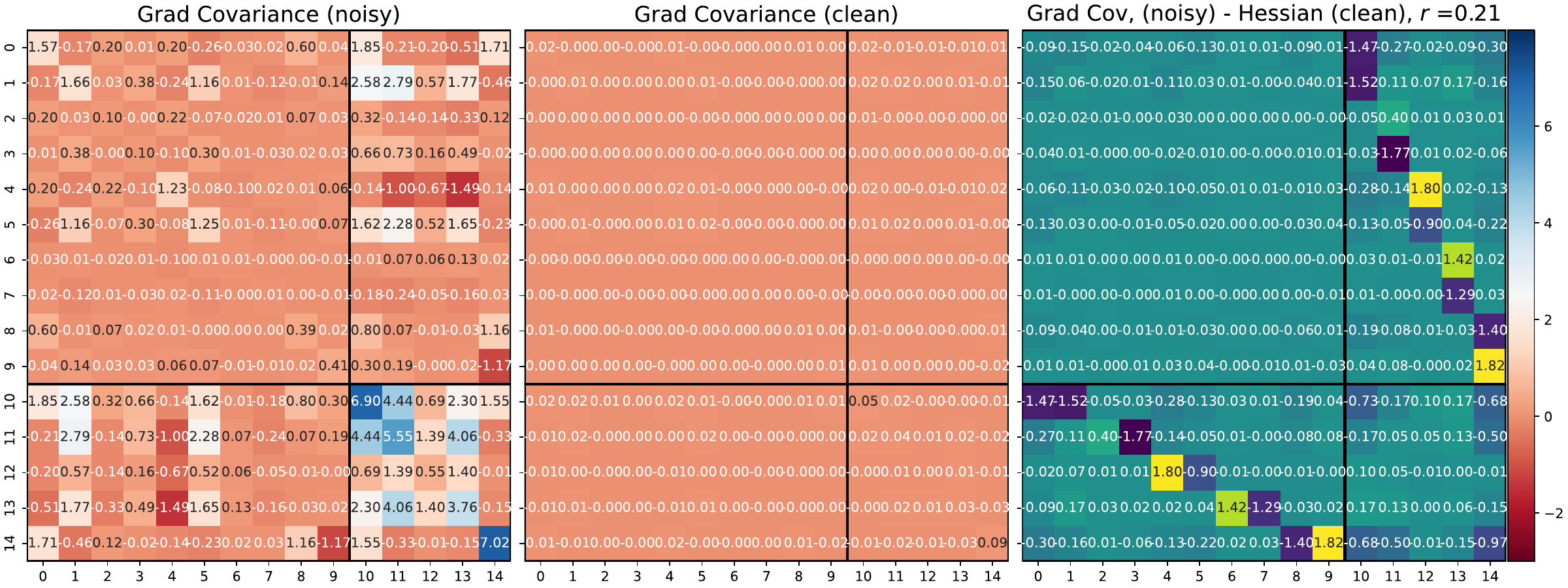}
    \caption{\textbf{Comparisons of Hessians and gradient covariance matrices computed with clean and noisy targets.} \textit{Top}: True Hessians calculated with noisy and clean targets and their difference. \textit{Middle}: Covariance matrices of noisy and clean gradients and the difference with the Hessian with noisy targets. \textit{Bottom}: Covariance matrices of noisy and clean gradients and the difference with the Hessian with clean targets. The lines show diagonal blocks of Hessians corresponding to the individual layers of size ten and five.}
    \label{fig:h_diff}
\end{figure}

The diagonal blocks of the Hessian for an MLP network with ReLU nonlinearities are independent of the targets in the case of mean-squared error loss functions. The significance of this observation lies in the fact that, even when gradients are noisy, our Hessian estimates remain accurate for the diagonal blocks, as the Hessians of these blocks are independent of the target. The diagonal blocks of the Hessian are the sub-matrices consisting of the second derivatives with respect to each group of variables individually, where the groups of variables correspond to layer parameters. We empirically demonstrate the independence of the blocks on the targets using generated data. 

\textbf{Data generation} We generated a dataset of 10000 samples by drawing two features from a Gaussian distribution with zero mean and standard deviation equal to 5. To obtain the targets, we pass the generated features through a hidden layer of five neurons and an output layer, with no bias term in either the hidden or output layers.

In this experiment, we run the comparison with a batch size equal to 64 (and the noise variance equal to 8 correspondingly). In the evaluation, we used an MLP of the same size as in the data generation, with the true weights distorted by a random vector with the norm equal to 0.1. Fig.~\ref{fig:h_diff} demonstrates Hessian matrices obtained with clean and noisy gradients, the gradient covariance matrices of clean and noisy gradients and their difference with the true Hessians.

\subsubsection{Hessian approximation on generated data with different network sizes}
\label{sec:mlp_gen_diff_size_hessian}
We further tested the quality of the estimate of the Hessians by the covariance matrix of noisy gradients during model training with MLPs of varying sizes. 

In the experiment, we generated data by drawing 10K samples with 10 features from a Gaussian distribution with a zero mean and a standard deviation of five, and then passing them through an MLP (similar to the previous section). The resulting values were perturbed by adding Gaussian noise with a standard deviation equal to 0.1 to obtain the final targets. We tested the approximation of the true Hessian using the covariance matrices of the gradients obtained from the calculated targets and the targets with additional noise added, with variance equal to the batch size. We used a batch size of 64 with the Adam optimiser and a learning rate of 0.01. For each dataset, 10\% of the data is reserved as a test set.

\begin{figure}[!t]
    \centering
    \includegraphics[width=0.99\linewidth, trim={1 1 1 1}, clip]{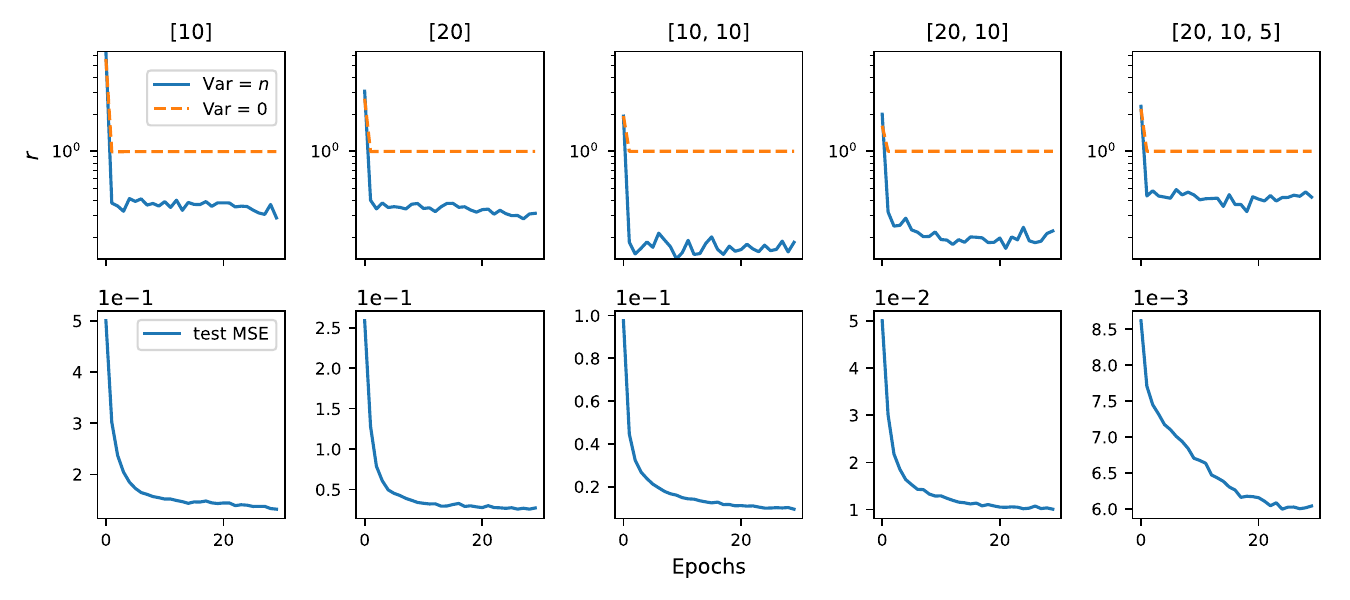}
    \caption{\textbf{Comparisons of Hessians estimated by the gradient covariance for MLPs.} The size of the network is indicated above. \textit{Top}: The relative operator norm $r$ for clean and noisy gradients. \textit{Bottom}: Mean absolute error on the hold-out test set as a measure for convergence. All presented results are averaged across 10 random seeds.}
    \label{fig:mlp_gen}
\end{figure}

We tested the MLPs that have the hidden layers of size (a)[10], (b) [20], (c) [10, 10], (d) [20, 10], (e) [20, 10, 5] and an output layer. The initial weights for the model are set to the true parameters used for data generation, but they are distorted by a random vector with a norm of three. The results are summarised in Fig.~\ref{fig:mlp_gen}. The obtained Hessian approximations are informative in a neighbourhood of the correct solution, and the results hold for the models of varying size.

\subsection{Gradient covariance estimate as a pre-conditioner}
\label{sec:pc_results}

In this section, we provide training details on the runs with the gradient covariance estimate as a preconditioner and provide an additional result with stochastic parameter updates performed after every 50 batches.

\subsubsection{Training details}

In this experiment, we run all models with a warm-up stage and then the main training stage. The warmup stage is performed with Adam optimiser, updates with noisy mean gradients and no preconditioning for all configurations tested in the main stage. We updated the model parameters after every 50 batches in the warmup stage. The main stage is presented in the results and was performed with gradient descent (GD) and Adam optimisers, the updates with clean and noisy gradients and preconditioning. The results in the main text correspond to the setting when model weights are updated once per epoch (full batch training). Additionally, we showed that preconditioning and weight updates with the noisy gradients are possible with a stochastic version of the algorithms in the main stage as well. 

\textbf{Full batch training} We used the following number of warmup epochs, warmup learning rates, number of epochs and learning rates in the main stage: for the bike sharing data, 10 epochs with a learning rate of 0.01 (warmup), 500 epochs with a learning rate of 0.003 and 500 epochs with 0.0007 (main), for the California housing data, 10 epochs with a learning rate of 0.01 (warmup), 3000 epochs with a learning rate of 5e-4 (main), for the wine quality data, 100 epochs with a learning rate of 0.01 (warmup), 2500 epochs with a learning rate of 8e-4 and 2500 epochs with a learning rate of 2e-4 (main), and for wave energy converters, 10 epochs with a learning rate of 0.001 (warmup), 400 epochs with a learning rate of 7e-4 and 400 epochs with a learning rate of 7e-5 (main). The batch size of 64 is used for all datasets considered in the experiment.

\textbf{Updates after every 50 batches (stochastic)} The parameters for the warmup stage are the same as in the full batch training. For the main training stage, we used the following number of epochs and learning rates: for the bike sharing data, 500 epochs with a learning rate of 5e-4 and 500 epochs with 7e-5, for the California housing data, 3000 epochs with a learning rate of 5e-4, for the wine quality data, 2500 epochs with a learning rate of 2e-4 and 2500 epochs with a learning rate of 5e-5, and for wave energy converters, 400 epochs with a learning rate of 5e-5 and 400 epochs with a learning rate of 8e-6. The batch size of 64 is used for all datasets.

\subsubsection{Stochastic optimisation results}

\begin{figure}[!t]
    \centering
    \includegraphics[width=1.0\linewidth, trim={0 70 0 0}, clip]{figures/noisy_opt_comparison.pdf}
    \includegraphics[width=1.0\linewidth, trim={0 0 0 24}, clip]{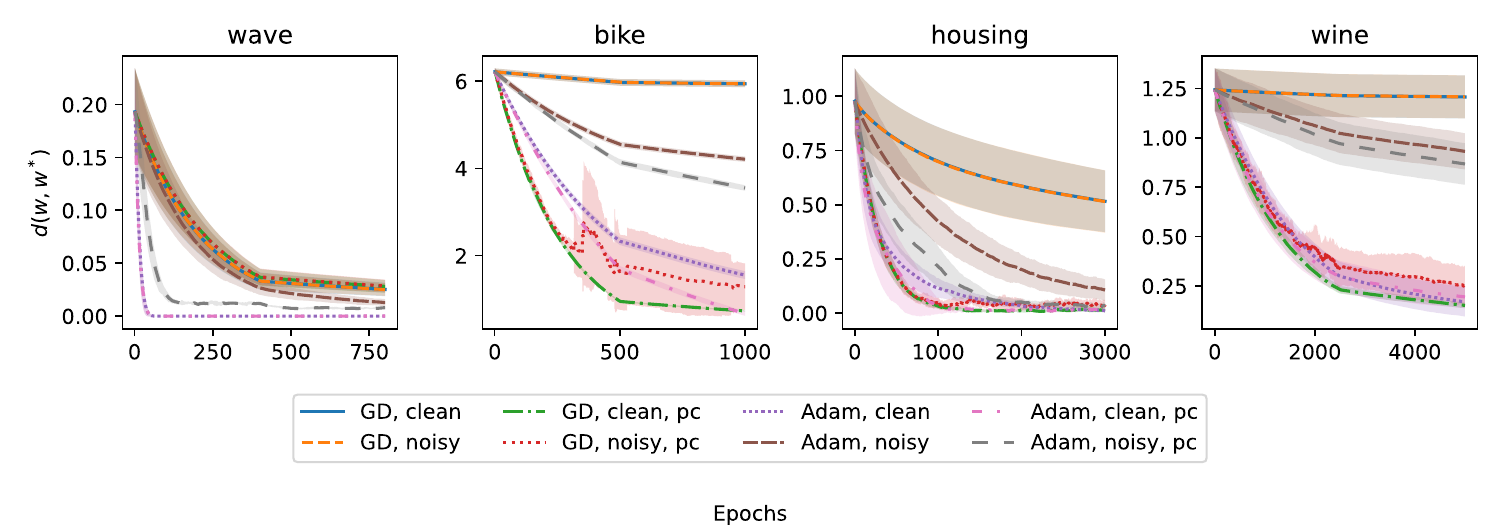}
    \caption{\textbf{Comparisons of optimisation algorithms with clean and noisy gradient updates and the gradient covariance as a preconditioner.} \textit{Top}: full batch updates (the same as in the main text). \textit{Bottom}: stochastic updates with the mean gradients computed from 50 batches.}
    \label{fig:pc_sgd}
\end{figure}

In this section, we present additional results for the preconditioning with a Hessian estimate obtained from noisy gradients. We tested full batch training (covered in the main text) and the stochastic version of the optimisation algorithms. In the stochastic version, the batch gradients are averaged across every 50 batches to perform the update, while the preconditioner is updated once per epoch. The results are presented in Fig.~\ref{fig:pc_sgd}.

\vfill

\end{document}